\documentclass[10pt]{article}

\usepackage[accepted]{rlj}

\usepackage{amssymb}
\usepackage{mathtools}
\usepackage{mathrsfs}
\usepackage{graphicx}
\usepackage{subcaption}
\usepackage[space]{grffile}
\usepackage{url}
\usepackage{lipsum}

\usepackage{pgfplots}
\usepgfplotslibrary{fillbetween}
\usepgfplotslibrary{groupplots}
\pgfplotsset{compat=1.18}
\usepackage{booktabs}
\usepackage{multirow}

\newcommand{\eg}{e\/.\/g\/.,\/~}

\title{Biased Dreams: Limitations to Epistemic Uncertainty Quantification in Latent Dynamics Models}
\setrunningtitle{Biased Dreams: Limitations to Epistemic Uncertainty Quantification in Latent Dynamics Models}

\author{Julia Berger\textsuperscript{1, $^*$}, Bernd Frauenknecht\textsuperscript{1, $^*$}, Sebastian Trimpe\textsuperscript{1}, Bastian Leibe\textsuperscript{1}}
\emails{\{berger,leibe\}@vision.rwth-aachen.de \\ \{bernd.frauenknecht,sebastian.trimpe\}@dsme.rwth-aachen.de}
\affiliations{
$^{1}$\textbf{RWTH Aachen University}
\par
$^*$ Equal contribution
}

\contribution{
    \textit{Latent Attractor Behavior.} 
    We provide empirical evidence that Recurrent State Space Models (RSSMs) can exhibit an attractor behavior, where latent transitions are biased toward well-supported latent regions that may diverge from the true environment dynamics.
    }
    {
    Proprioceptive dynamics models typically accumulate prediction errors during prolonged rollouts, leading to increasingly unstable predictions.
    }

\contribution{
    \textit{Limits of Epistemic Uncertainty.} 
    Although ensemble disagreement reliably captures \textit{local} epistemic uncertainty, we demonstrate that the attractor behavior can decouple uncertainty estimates from \textit{global} trajectory-level model error in RSSM latent rollouts.
    }
    {
    Ensemble-based epistemic uncertainty quantification is well-established for proprioceptive dynamics models and is often transferred directly to latent dynamics models without a careful analysis of its reliability.
    }

\contribution{
    \textit{Return Overestimation in Latent Rollouts.}
    We observe that the attractor behavior can result in overly optimistic return predictions in RSSM latent rollouts when attractor regions coincide with high-reward behaviors.
    }
    {
    This high-reward bias during imagination may distort policy learning by reinforcing overconfident behaviors that do not correspond to true environment outcomes.
    }


\summary{
Model-based reinforcement learning distinguishes between dynamics models operating on proprioceptive states and latent dynamics models typically operating on high-dimensional image observations. Among the latter, Dreamer's Recurrent State Space Model (RSSM) has emerged as a dominant architecture. While ensemble-based epistemic uncertainty has proven effective in proprioceptive dynamics for mitigating model exploitation, guiding exploration, or promoting caution, its behavior in latent dynamics remains largely unexplored. Our experiments reveal that although ensemble disagreement captures \textit{local} epistemic uncertainty, it does not reliably reflect \textit{global} compounding model error accumulated over prolonged RSSM latent rollouts. We provide evidence for an attractor behavior that draws rollouts toward well-supported latent regions, where uncertainty diminishes despite increasing discrepancies from the true environment dynamics. This can cause the model to overestimate returns when attractor regions correspond to high-reward behaviors. Our findings reveal a structural limitation of epistemic uncertainty estimation in RSSMs and challenge the assumption that epistemic uncertainty estimation transfers directly from proprioceptive to latent dynamics.
}

\begin{document}

\makeCover
\maketitle

\begin{abstract}
Model-based reinforcement learning distinguishes between dynamics models operating on proprioceptive states and latent dynamics models typically operating on high-dimensional image observations. Among the latter, Dreamer's Recurrent State Space Model (RSSM) has emerged as a dominant architecture. While ensemble-based epistemic uncertainty has proven effective in proprioceptive dynamics for mitigating model exploitation, guiding exploration, or promoting caution, its behavior in latent dynamics remains largely unexplored. Our experiments reveal that although ensemble disagreement captures \textit{local} epistemic uncertainty, it does not reliably reflect \textit{global} compounding model error accumulated over prolonged RSSM latent rollouts. We provide evidence for an attractor behavior that draws rollouts toward well-supported latent regions, where uncertainty diminishes despite increasing discrepancies from the true environment dynamics. This can cause the model to overestimate returns when attractor regions correspond to high-reward behaviors. Our findings reveal a structural limitation of epistemic uncertainty estimation in RSSMs and challenge the assumption that epistemic uncertainty estimation transfers directly from proprioceptive to latent dynamics.
\end{abstract}

%%%%%%%%%%%%%%%%%%%%%%%%%%%%%%%%%%%%%%%%%%%%%%%%%%%%%%%%%%%%%%%%
%% Sections
%%%%%%%%%%%%%%%%%%%%%%%%%%%%%%%%%%%%%%%%%%%%%%%%%%%%%%%%%%%%%%%%
%%%%%%%%%%%%%%%%%%%%%%%%%%%%%%%%%%%%%%%%%%%%%%%%%%%%%%%%%%%%%%%%
%% Section: Introduction
%%%%%%%%%%%%%%%%%%%%%%%%%%%%%%%%%%%%%%%%%%%%%%%%%%%%%%%%%%%%%%%%
\section{Introduction}
\label{sec:introduction}

Reinforcement Learning (RL) provides a powerful framework for solving complex control problems~\citep{berner2019dota,degrave2022magnetic}. Model-Based Reinforcement Learning (MBRL) addresses the sample inefficiency of standard RL by learning a parametric model of the environment dynamics to generate artificial interaction data for policy learning. Within the MBRL literature, we distinguish between \textit{proprioceptive dynamics models}~\citep{chua2018deep}, which operate on proprioceptive inputs, and \textit{latent dynamics models}, which typically learn compact state representations from high-dimensional image observations~\citep{finn2017deep,ha2018world}. Among latent approaches, Dreamer~\citep{hafner2019dreamerv1,hafner2020dreamerv2,hafner2023dreamerv3} has established the Recurrent State Space Model (RSSM)~\citep{hafner2019planet} as a dominant architecture for vision-based MBRL. While alternative latent dynamics have been proposed~\citep{hansen2023tdmpc2,zhou2024dino,assran2025v}, RSSMs remain a widely employed and computationally efficient baseline.

\begin{figure}[htb!]
  \centering
  \begin{minipage}[b]{0.485\textwidth}
    \centering
    RSSM\\
    \includegraphics[width=\textwidth]{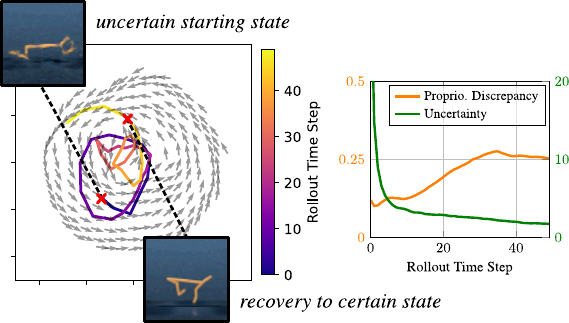}
  \end{minipage}
  \hfill
  \begin{minipage}[b]{0.485\textwidth}
    \centering
    PE\\
    \includegraphics[width=\textwidth]{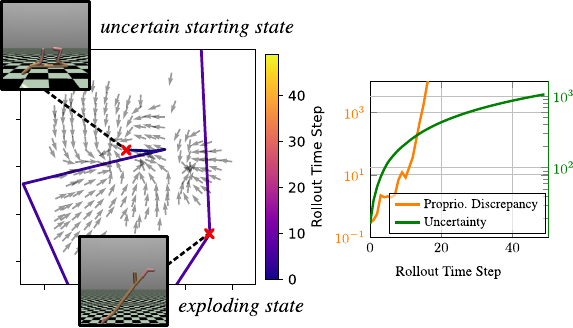}
  \end{minipage}
  \caption{Conceptual illustration of our key findings. For an RSSM and a proprioceptive probabilistic ensemble (PE) model, the left subplots show PCA-embedded transition dynamics with an exemplary rollout initialized from an out-of-distribution state. The right subplots show proprioceptive discrepancy, which is the mismatch between predicted and true proprioceptive states under the same action sequence, and ensemble-based epistemic uncertainty. While the PE rollout becomes unstable with increasing prediction error and high uncertainty, the RSSM rollout converges to familiar latent regions, where uncertainty diminishes despite elevated proprioceptive discrepancy.}
\label{fig:teaser}
\end{figure}

A central question in MBRL is whether model-based rollouts provide reliable predictions for decision-making, as compounding model error can deteriorate policy learning~\citep{janner2019mbpo,yu2020mopo}. Ensemble-based epistemic uncertainty quantification is widely leveraged to identify unreliable model predictions~\citep{lakshminarayanan2017simple}. While this \textit{local} epistemic uncertainty has been shown to be informative about \textit{global} trajectory-level model error accumulated during rollouts in proprioceptive dynamics models~\citep{janner2019mbpo,yu2020mopo,frauenknecht2024macura}, this relationship is often implicitly assumed to transfer to latent dynamics~\citep{wang2020dmve,zhu2020bird,seyde2020love,filos2022modelvalueinconsistency,seo2025unisafe,liu2026perceiving}. In this work, we study whether this relationship transfers to RSSMs and provide, to the best of our knowledge, the first systematic analysis of epistemic uncertainty quantification in this setting. We summarize our findings below, conceptually illustrated in Fig.~\ref{fig:teaser}:
\begin{enumerate}[label=\textit{(\roman*)}]
    \item RSSM latent rollouts can exhibit an attractor behavior, where transitions are biased toward well-supported regions in latent space that may diverge from the true environment dynamics.
    \item The attractor behavior can mask discrepancies between latent and environment dynamics, causing epistemic uncertainty estimates to become decoupled from compounding model error.
    \item Attractor regions may coincide with high-reward behaviors, causing RSSM latent rollouts to predict overly optimistic returns.
\end{enumerate}

%%%%%%%%%%%%%%%%%%%%%%%%%%%%%%%%%%%%%%%%%%%%%%%%%%%%%%%%%%%%%%%%
%% Section: Background
%%%%%%%%%%%%%%%%%%%%%%%%%%%%%%%%%%%%%%%%%%%%%%%%%%%%%%%%%%%%%%%%
\section{Background}
\label{sec:background}

In the following, we introduce the foundations of Reinforcement Learning (RL), Recurrent State Space Models (RSSMs), and uncertainty quantification.

%%%%%%%%%%%%%%%%%%%%%%%%%%%%%%%%%%%%%%%%%%%%%%%%%%%%%%%%%%%%%%%%

\subsection{Reinforcement Learning under Full and Partial Observability}
\label{subsec:back_rl}

Reinforcement Learning (RL) addresses sequential decision-making via agent–environment interaction. For an observable proprioceptive environment state $\mathbf{s}_t \in \mathcal{S}$, the control problem can be formulated as a Markov Decision Process (MDP)~\citep{andrew2018reinforcement}. At each step, the agent selects an action $\mathbf{a}_{t} \sim \pi(\cdot \mid \mathbf{s}_{t})$, after which the environment provides the next state and reward $\mathbf{s}_{t+1}, r_{t} \sim p(\cdot, \cdot\mid \mathbf{s}_{t}, \mathbf{a}_{t})$. The objective is to learn a policy maximizing the expected discounted return $\mathbb{E}_{\pi}\left( \sum_{t=1}^\infty \gamma^t r_t\right)$ with discount factor $0<\gamma<1$. In Model-Based Reinforcement Learning (MBRL), a learned model $\tilde p(\mathbf{s}_{t+1} \mid \mathbf{s}_{t}, \mathbf{a}_{t})$ approximates the unknown environment dynamics and enables planning or simulated interactions. Early approaches employ Gaussian processes~\citep{deisenroth2011pilco}, local linear models~\citep{levine2013guided}, and deterministic neural networks~\citep{williams2017information, nagabandi2018neural}. Recent successful MBRL methods on proprioceptive inputs~\citep{chua2018deep, janner2019mbpo,yu2020mopo} typically choose the probabilistic ensemble (PE)~\citep{lakshminarayanan2017simple} model architecture.

In visual control settings, the underlying state $\mathbf{s}_t$ is not directly observable, and the problem is therefore formulated as a Partially Observable Markov Decision Process (POMDP)~\citep{andrew2018reinforcement}. At each step, the agent selects an action $\mathbf{a}_t \sim \pi(\cdot \mid \mathbf{o}_{\leq t}, \mathbf{a}_{< t})$ conditioned on past observations and actions, after which the environment provides the next observation and reward $\mathbf{o}_{t+1}, r_{t+1} \sim p(\cdot, \cdot \mid \mathbf{o}_{< t}, \mathbf{a}_{< t})$. The agent typically maintains a belief state summarizing the history of observations and actions used in decision-making. Because modeling dynamics directly in high-dimensional observation space is impractical, many approaches learn a compact latent representation that summarizes this history and enables predictive transitions~\citep{finn2017deep,ha2018world}. Such models are commonly referred to as latent dynamics models.

%%%%%%%%%%%%%%%%%%%%%%%%%%%%%%%%%%%%%%%%%%%%%%%%%%%%%%%%%%%%%%%%

\subsection{Latent Dynamics Modeling via Recurrent State Space Models}
\label{sec:background_latent_dynamics}

The Recurrent State Space Model (RSSM)~\citep{hafner2019planet} is a latent dynamics model that represents a belief state using stochastic state $\mathbf{z}_t$ and deterministic recurrent state $\mathbf{h}_t$. It consists of
\begin{align}
    \text{Representation model:}\quad &q(\mathbf{z}_t \mid \mathbf{z}_{t-1}, \mathbf{a}_{t-1}, \mathbf{o}_{t}), \label{eq:repr_model}\\
    \text{Observation model:}\quad &p(\mathbf{o}_t \mid \mathbf{z}_{t}),\\
    \text{Transition model:}\quad &p(\mathbf{z}_{t} \mid \mathbf{z}_{t-1}, \mathbf{a}_{t-1}),\label{eq:latent_trans_mod}\\
    \text{Reward model:}\quad &p(r_t \mid \mathbf{z}_{t}).
    \label{eq:rew_model}
\end{align}
For simplicity, we omit explicit dependencies $\mathbf{z}_{< t-1}, \mathbf{a}_{< t-1}$ induced by the recurrence of $\mathbf{h}_t$. All distributions are modeled as multivariate Gaussians with diagonal covariance.

We define two types of latent rollouts starting at some initial state $\mathbf{z}_0$. \textit{Prior} rollouts $\tau^{\text{prior}}$ are generated solely by the transition model (Eq.~\eqref{eq:latent_trans_mod}) and \textit{posterior-informed} rollouts $\tau^{\text{post-inf}}$ perform one-step transition model predictions (Eq.~\eqref{eq:latent_trans_mod}) initialized from posterior latent states $\mathbf{z}_{t-1} \sim q(\mathbf{z}_{t-1} \mid \cdot)$:
\begin{align}
    \tau^{\text{prior}} &= \{(\mathbf{z}_t, \mathbf{a}_t)\}_{t=0}^{T-1}, \quad \mathbf{z}_{t} \sim p(\mathbf{z}_t \mid \mathbf{z}_{t-1},\mathbf{a}_{t-1});
    \label{eq:prior_rollouts}\\
    \tau^{\text{post-inf}} &= \{(\mathbf{z}_t, \mathbf{a}_t)\}_{t=0}^{T-1}, \quad \mathbf{z}_{t} \sim p(\mathbf{z}_t \mid \mathbf{z}_{t-1},\mathbf{a}_{t-1}) \quad\text{and}\quad \mathbf{z}_{t-1} \sim q(\mathbf{z}_{t-1} \mid \cdot).
    \label{eq:post_inf_rollouts}
\end{align}

RSSM training involves maximizing the Evidence Lower Bound (ELBO)~\citep{hafner2019planet}
\begin{align}
    \log &\,p(\mathbf{o}_{\leq T} \mid \mathbf{a}_{\leq T}) 
        \geq \sum_t \Big( 
        \mathbb{E}_{q(\mathbf{z}_t \mid \mathbf{o}_{\leq t}, \mathbf{a}_{< t})}
        \left[ \log p(\mathbf{o}_t \mid \mathbf{z}_{t}) + \log p(r_t \mid \mathbf{z}_{t})\right] \nonumber \\ 
        &- \mathbb{E}_{q(\mathbf{z}_{t-1} \mid \mathbf{o}_{\leq t-1}, \mathbf{a}_{< t-1})}\left[ D_{\text{KL}}\left(q(\mathbf{z}_t \mid \mathbf{z}_{t-1}, \mathbf{a}_{t-1}, \mathbf{o}_{t}) \,\middle\|\,p(\mathbf{z}_{t} \mid \mathbf{z}_{t-1}, \mathbf{a}_{t-1})\right)\right]
    \Big).
    \label{eq:rssm_elbo}
\end{align}
DreamerV2/V3~\citep{hafner2020dreamerv2,hafner2023dreamerv3} extend Dreamer~\citep{hafner2019dreamerv1} with stability improvements and a categorical RSSM variant (Cat-RSSM), where the stochastic latent state is modeled using discrete categorical variables.

%%%%%%%%%%%%%%%%%%%%%%%%%%%%%%%%%%%%%%%%%%%%%%%%%%%%%%%%%%%%%%%%

\subsection{Uncertainty in Model-Based Reinforcement Learning}
\label{subsec:background_latent_epi}

Uncertainty is commonly divided into an aleatoric component, describing irreducible environment stochasticity, and an epistemic component, arising from limited data or imperfect learning. In proprioceptive dynamics models, the PE~\citep{lakshminarayanan2017simple} represents aleatoric uncertainty as a Gaussian distribution over the next state, while epistemic uncertainty is captured by disagreement among $M$ ensemble members $\{p_{i}(\mathbf{s}_{t+1} \mid \mathbf{s}_{t}, \mathbf{a}_{t})\}_{i=1}^M$. Consequently, aligned ensemble predictions indicate a reliable approximation of environment dynamics, assuming sufficient representational capacity of the model~\citep{yu2020mopo,frauenknecht2024macura}.

Similarly, latent dynamics models often use ensembles of latent transition predictors described as $\{g_{i}(\mathbf{z}_{t} \mid \mathbf{z}_{t-1}, \mathbf{a}_{t-1})\}_{i=1}^M$ to measure disagreement in latent space~\citep{sekar2020plan2explore}. Alignment of ensemble predictions for a given pair $(\mathbf{z}_{t-1}, \mathbf{a}_{t-1})$ indicates that the latent transition model (Eq.~\eqref{eq:latent_trans_mod}) is well-approximated~\citep{lakshminarayanan2017simple}, which is typically interpreted as evidence that the latent dynamics reflect true environment dynamics~\citep{sekar2020plan2explore,seo2025unisafe}.
%%%%%%%%%%%%%%%%%%%%%%%%%%%%%%%%%%%%%%%%%%%%%%%%%%%%%%%%%%%%%%%%
%% Section: Related Work
%%%%%%%%%%%%%%%%%%%%%%%%%%%%%%%%%%%%%%%%%%%%%%%%%%%%%%%%%%%%%%%%
\section{Related Work}
\label{sec:related_work}

\paragraph{Proprioceptive Dynamics Models.} Proprioceptive dynamics models are commonly learned as PE-based predictors over relatively low-dimensional state spaces~\citep{lakshminarayanan2017simple}. Epistemic uncertainty in these models is used to constrain model exploitation~\citep{yu2020mopo,buckman2018sample,frauenknecht2024macura,frauenknecht2025infoprop}, guide exploration~\citep{sancaktar2022curious,NEURIPS2023_77b5aaf2}, or promote caution~\citep{yu2023safe,vignola2026sampling,frauenknecht2026uncertainty}.

\paragraph{Latent Dynamics Models.} In vision-based RL, latent dynamics models learn compact representations from high-dimensional image observations. Approaches include reconstruction-based models~\citep{finn2017deep,ha2018world,zhang2019solar}, self-supervised representation learning methods~\citep{pathak2017curiosity,schwarzer2020data,hansen2022tdmpc,zhou2024dino}, and sequence-based architectures~\citep{chen2022transdreamer,micheli2022transformers}. Among these, the RSSM~\citep{hafner2019planet,hafner2019dreamerv1,hafner2020dreamerv2,hafner2023dreamerv3} is a widely used and efficient baseline. Uncertainty-aware objectives in latent dynamics models fall into two main categories: explicit epistemic uncertainty estimates used for robustness~\citep{wang2020dmve,zhu2020bird,seyde2020love,filos2022modelvalueinconsistency,liu2026perceiving}, exploration~\citep{sekar2020plan2explore,sancaktar2025sensei}, or safety~\citep{seo2025unisafe,nakamura2025generalizing}; and prediction-based intrinsic signals such as curiosity~\citep{guo2022byol,jarrett2022curiosity}. While differing in formulation, both objectives rely on the assumption that uncertainty-related signals in latent space reflect long-term transition reliability. In this work, we revisit this assumption for RSSMs using ensemble-based epistemic uncertainty estimates, which closely align with uncertainty modeling practices in proprioceptive dynamics models.
%%%%%%%%%%%%%%%%%%%%%%%%%%%%%%%%%%%%%%%%%%%%%%%%%%%%%%%%%%%%%%%%
%% Section: Problem Statement
%%%%%%%%%%%%%%%%%%%%%%%%%%%%%%%%%%%%%%%%%%%%%%%%%%%%%%%%%%%%%%%%
\section{Problem Statement}
\label{sec:problem_statement}

Epistemic uncertainty quantification via ensemble disagreement is well-established in proprioceptive dynamics models~\citep{yu2020mopo,frauenknecht2024macura,frauenknecht2025infoprop} and has increasingly been adopted in latent dynamics models, in particular RSSM-based approaches~\citep{seyde2020love,sekar2020plan2explore,seo2025unisafe}, where ensemble disagreement between transition predictors~\citep{sekar2020plan2explore} serves as the latent counterpart of proprioceptive PE-based uncertainty. This correspondence makes RSSMs a representative architecture to study whether uncertainty estimation principles transfer from proprioceptive to latent dynamics. In particular, employing ensemble disagreement in latent dynamics implicitly assumes that \textit{local} transition uncertainty remains informative about \textit{global} trajectory-level model error during prolonged rollouts, as observed in proprioceptive dynamics models~\citep{janner2019mbpo}. In this work, we empirically revisit this assumption via three questions:
\begin{enumerate}[label=\textit{(\roman*)}]
    \item \label{prob1} How do RSSM rollouts evolve under in- and out-of-distribution settings compared to proprioceptive dynamics models?
    \item \label{prob2} Does \textit{local} ensemble-based epistemic uncertainty in RSSM latent dynamics remain informative about \textit{global} trajectory-level deviations from true environment dynamics?
    \item \label{prob3} How does the RSSM latent behavior affect reward prediction and therefore policy learning?
\end{enumerate}
Our analysis reveals a failure mode in RSSM latent dynamics where epistemic uncertainty decouples from compounding model error during prolonged rollouts, leading to reward overestimation.

%%%%%%%%%%%%%%%%%%%%%%%%%%%%%%%%%%%%%%%%%%%%%%%%%%%%%%%%%%%%%%%%
%% Section: Problem Evaluation
%%%%%%%%%%%%%%%%%%%%%%%%%%%%%%%%%%%%%%%%%%%%%%%%%%%%%%%%%%%%%%%%
\section{Problem Evaluation}

We empirically investigate the above questions using RSSM~\citep{hafner2019planet} and Cat-RSSM~\citep{hafner2020dreamerv2,hafner2023dreamerv3}. Section~\ref{sec:attr_eval} analyzes latent and proprioceptive rollout behavior under a distribution shift, Section~\ref{sec:phys_discr} relates ensemble uncertainty to proprioceptive discrepancy, and Section~\ref{sec:reward_eval} evaluates the impact of latent-environment misalignment on reward prediction.

\paragraph{Experimental Setup.} Experiments are conduced on four DMC Suite tasks~\citep{tassa2018dmc}: Cartpole Swingup, Cheetah Run, Hopper Hop, and Walker Run. For each task, we perform $S=5$ training runs with different random seeds. The RSSM implementation follows~\citet{becker2022vrkn}, while the Cat-RSSM implementation follows~\citet{becker2023coral}, isolating the categorical latent space by omitting other DreamerV2/V3 modifications. All runs are trained for 1 Mio. environment steps. Additional details of our implementation\footnote{\url{https://github.com/jberger999/biased-dreams}} are provided in Appendix~\ref{sec:supp_exp}.

\paragraph{Uncertainty Estimation.} We train an external ensemble of $M=5$ one-step latent transition predictors $\{g_{i}(\mathbf{z}_{t} \mid \mathbf{z}_{t-1}, \mathbf{a}_{t-1})\}_{i=1}^M$, each optimized via the negative Gaussian log-likelihood of the next latent state predicted by the representation model (Eq.~\ref{eq:repr_model}). Following~\citet{frauenknecht2024macura}, epistemic uncertainty is computed as the ensemble disagreement over predicted next latent state distributions using Geometric Jensen-Shannon divergence~\citep{nielsen2019jensen}.

\paragraph{Random Rollouts Collection.} For each trained run, we collect three types of rollouts from $N=1000$ random initial states with rollout length $T=50$ using the trained greedy policy: (1) \textit{posterior-informed} rollouts (Eq.~\eqref{eq:post_inf_rollouts}); (2) \textit{closed-loop prior} rollouts (Eq.~\eqref{eq:prior_rollouts}), initialized from posterior-informed states; and (3) \textit{open-loop prior} rollouts, using actions from posterior-informed rollouts. Closed-loop refers to rollouts where actions are selected by a reactive policy based on the current latent states, whereas open-loop rollouts follow a predefined action sequence without feedback from the policy. Posterior-informed rollouts represent the most informed rollouts the model can produce, leveraging observation-updated latent states and closed-loop actions while decoupling transition dynamics from observation noise. Closed-loop prior rollouts correspond to latent imagination used during Dreamer's policy learning, whereas open-loop prior rollouts isolate the effects of model stochasticity and error accumulation under a fixed action sequence.

\paragraph{ID and OOD Initial States.} To analyze model behavior under distribution shifts, we select in-distribution (ID) and out-of-distribution (OOD) initial states from the post-training rollouts under the learned policy. We use ensemble disagreement over latent states and actions as a measure of \textit{local} uncertainty, which has been shown to reliably capture epistemic transition uncertainty under distribution shifts~\citep{lakshminarayanan2017simple,chua2018deep}. ID states are selected as the $K$ lowest-uncertainty states from the posterior-informed and closed-loop prior rollouts. OOD states are selected as the $K$ highest-uncertainty states from the posterior-informed and open-loop prior rollouts, where fixed actions allow model stochasticity and error accumulation to progressively induce a state-action mismatch. We use $K=10$ initial states per seed, task, and setup (ID/OOD).

%-------------------------------------------------------------

\subsection{Attractor Evaluation}
\label{sec:attr_eval}

To address \textit{Question}~\ref{prob1}, we analyze how ID and OOD initial states affect rollout behavior in proprioceptive and latent dynamics models. Under a distribution shift, proprioceptive dynamics models typically exhibit prediction instabilities due to compounding model error~\citep{janner2019mbpo}. We investigate whether RSSMs exhibit a similar behavior.

\paragraph{Visualization of Dynamical Behavior.} To avoid attributing model behavior to suboptimal training, we select the best-performing run for each task, which has the highest mean evaluation return over the training. We embed its posterior-informed and closed-loop prior rollouts into a 2D PCA space of normalized deterministic features $\mathbf{f}_t := (\mathbf{h}_t,\mathbf{z}^{\text{m}}_t)$, where $\mathbf{z}^{\text{m}}_t$ denotes the mean stochastic state in RSSM and the categorical mode in Cat-RSSM. We construct a vector field by binning one-step displacements $\hat{\mathbf{f}}_{t} - \hat{\mathbf{f}}_{t-1}$, where $\text{PCA}(\mathbf{f}_{t})=\hat{\mathbf{f}}_{t}$, and overlay closed-loop rollouts initialized from ID states with lowest and OOD states with the highest mean uncertainty over the rollout horizon.

\paragraph{Proprioceptive Dynamics.} We additionally analyze PE dynamics from Infoprop~\citep{frauenknecht2025infoprop} in the MuJoCo HalfCheetah environment~\citep{todorov2012mujoco}, applying the same PCA procedure to proprioceptive states. ID and OOD rollouts are closed-loop rollouts with the lowest and highest mean uncertainty over the rollout horizon, respectively, selected from random initial states.

\begin{figure}[htbp]
    \centering
    \setlength{\tabcolsep}{0.5pt}
    \begin{tabular}{cc}
        RSSM & PE \\
        \hspace{1cm} \makebox[0pt][c]{\footnotesize ID} \hspace{2.5cm} \makebox[0pt][c]{\footnotesize OOD} \hspace{1.5cm}
        & \hspace{1cm} \makebox[0pt][c]{\footnotesize ID} \hspace{2.5cm} \makebox[0pt][c]{\footnotesize OOD} \hspace{1.5cm} \\
        \includegraphics[width=0.49\linewidth]{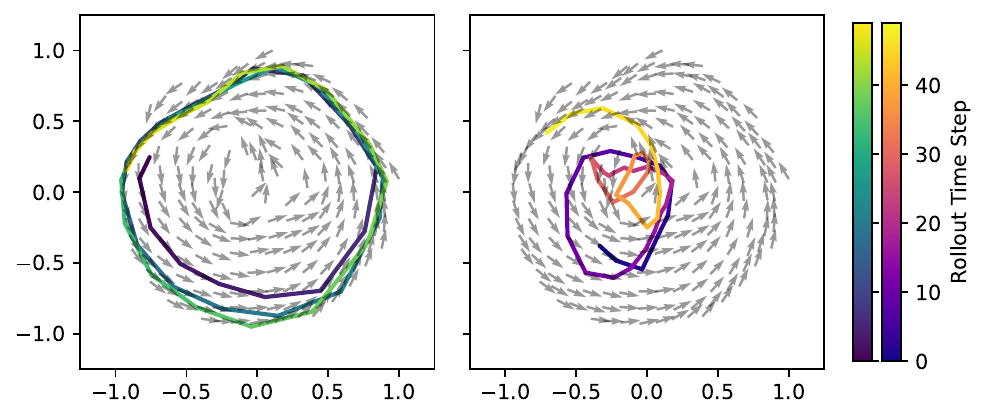} &
        \includegraphics[width=0.49\linewidth]{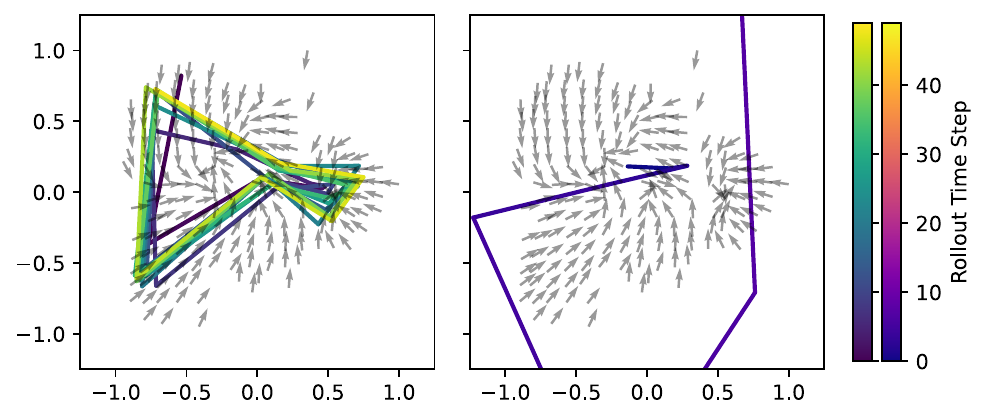} 
    \end{tabular}
    \caption{Attractor evaluation for RSSM and PE under ID and OOD initial states on Cheetah Run and Halfcheetah, respectively. Vector fields show one-step displacements in the 2D PCA spaces of latent and proprioceptive states. PE exhibits unstable OOD predictions, whereas RSSM is drawn toward dominant transition flows, indicating an attractor bias toward well-supported latent regions.}
    \label{fig:pca_plot}
\end{figure}

As visualized in Fig.~\ref{fig:pca_plot}, the PE exhibits unstable OOD predictions, reflecting compounding model error. In contrast, the RSSM OOD rollout converges toward dominant transition flows, indicating an attractor effect that biases rollouts toward well-supported latent regions. This behavior suggests that distribution shifts manifest differently in RSSMs and proprioceptive dynamics, as latent representations may constrain rollouts toward training-supported regions through transitions that are not necessarily physically accurate. A similar behavior can be observed across other DMC Suite tasks and for Cat-RSSM, as reported in Appendix~\ref{sec:supp_attractor_eval}. However, given the qualitative nature of this analysis, these findings should be interpreted as observations rather than definitive findings.

%-------------------------------------------------------------

\subsection{Proprioceptive Discrepancy}
\label{sec:phys_discr}

We evaluate proprioceptive discrepancy between RSSM latent rollouts and true environment behavior under ID/OOD initial states, and relate it to ensemble-based epistemic uncertainty, addressing \textit{Question}~\ref{prob2}. While ensemble disagreement reliably captures local epistemic uncertainty, we investigate whether this extends to \textit{global} trajectory-level prediction accuracy.

\paragraph{Proprioceptive State Decoder.} To compare latent predictions and true proprioceptive dynamics, we augment the RSSM with a proprioceptive state decoder (PD) $p(\mathbf{s}_t \mid \mathbf{z}_t)$, trained jointly via the ELBO (Eq.~\eqref{eq:rssm_elbo}) to predict transformed proprioceptive states excluding the agent’s $x$-position. Although the decoder shapes latent representations during training, proprioceptive reconstruction quality remains limited without an explicit structural objective~\citep{peper2025four,gupta2021towards}. As shown in Fig.~\ref{fig:rl_performance}, adding the PD has negligible impact on Dreamer's performance.

\begin{figure}[htb!]
\centering
% =====================================================
% DEFINITIONS
% =====================================================
\def\pathtocsvs{figs/rl_performance_plots}

\def\pathtofirst{\pathtocsvs/rssm/rl_performance-rssm-combined.csv}
\def\catpathtofirst{\pathtocsvs/cat_rssm/rl_performance-cat_rssm-combined.csv}

\begin{tikzpicture}

% ---------- COMMON STYLE ----------
\tikzstyle{every node}=[font=\footnotesize]
\pgfplotsset{
    myplot/.style={
        width=5.7cm,
        height=4.2cm,
        grid=major,
        xmin=0, xmax=1,
        ymin=0, ymax=1000,
        ylabel={Expected Return},
        every axis plot/.append style={line width=1.5pt},
        legend style={
            at={(0.01,0.99)},
            anchor=north west,
            thin,
            fill=white,
            fill opacity=0.6,
            draw opacity=1,
            text opacity=1,
            font=\small
        },
        label style={font=\small},
        tick label style={font=\small},
        title style={font=\small}
    }
}

\begin{groupplot}[
    group style={
        group size=2 by 1,
        horizontal sep=1.5cm,
        vertical sep=0.5cm
    },
    myplot
]

% =====================================================
% 11
% =====================================================
\nextgroupplot[
    title={RSSM}
]

% RSSM W/O PD
\addplot[forget plot,name path=vup, draw=none]
table[
    col sep=comma,
    x=step,
    y expr=\thisrow{upper}
]
{\pathtofirst};

\addplot[forget plot,name path=vlow, draw=none]
table[
    col sep=comma,
    x=step,
    y expr=\thisrow{lower}
]
{\pathtofirst};

\addplot[forget plot, orange, fill opacity=0.25] fill between[of=vup and vlow];

\addplot[orange]
table[
    col sep=comma,
    x=step,
    y=mean
]
{\pathtofirst};

\addlegendentry{w/o PD}

% RSSM W/ PD
\addplot[forget plot, name path=rup, draw=none]
table[
    col sep=comma,
    x=step,
    y expr=\thisrow{upper_phys_dec}
]
{\pathtofirst};

\addplot[forget plot, name path=rlow, draw=none]
table[
    col sep=comma,
    x=step,
    y expr=\thisrow{lower_phys_dec}
]
{\pathtofirst};

\addplot[forget plot, green!70!black, fill opacity=0.25] fill between[of=rup and rlow];

\addplot[green!70!black]
table[
    col sep=comma,
    x=step,
    y=mean_phys_dec
]
{\pathtofirst};

\addlegendentry{w/ PD}

% =====================================================
% 12
% =====================================================
\nextgroupplot[
    title={Cat-RSSM},
    ylabel={},
    yticklabels={}
]

% RSSM W/O PD
\addplot[forget plot, name path=vup2, draw=none]
table[
    col sep=comma,
    x=step,
    y expr=\thisrow{upper}
]
{\catpathtofirst};

\addplot[forget plot, name path=vlow2, draw=none]
table[
    col sep=comma,
    x=step,
    y expr=\thisrow{lower}
]
{\catpathtofirst};

\addplot[forget plot, magenta, fill opacity=0.25] fill between[of=vup2 and vlow2];

\addplot[magenta]
table[
    col sep=comma,
    x=step,
    y=mean
]
{\catpathtofirst};

\addlegendentry{w/o PD}

% RSSM W/ PD
\addplot[forget plot, name path=rup2, draw=none]
table[
    col sep=comma,
    x=step,
    y expr=\thisrow{upper_phys_dec}
]
{\catpathtofirst};

\addplot[forget plot, name path=rlow2, draw=none]
table[
    col sep=comma,
    x=step,
    y expr=\thisrow{lower_phys_dec}
]
{\catpathtofirst};

\addplot[forget plot, cyan, fill opacity=0.25] fill between[of=rup2 and rlow2];

\addplot[cyan]
table[
    col sep=comma,
    x=step,
    y=mean_phys_dec
]
{\catpathtofirst};

\addlegendentry{w/ PD}

\end{groupplot}

\end{tikzpicture}\\
{\small Environment Step ($\times 10^6$)}
\caption{Dreamer's performance with and without a proprioceptive state decoder (PD) for RSSM and Cat-RSSM across four DMC Suite tasks. The addition of a PD has negligible impact on policy learning, with both architectures showing comparable performance to their respective no-PD variants. Results are averaged over 5 seeds per setting; shaded areas indicate 95\% confidence intervals.}
\label{fig:rl_performance}
\end{figure}

\paragraph{Latent Dynamics Rollouts.} For each of the $K$ ID and OOD initial states, we evaluate $R$ closed-loop posterior-informed and $R$ closed-loop prior rollouts of length $T=50$. These rollout types provide complementary views of the learned dynamics: Closed-loop posterior-informed rollouts leverage observation-updated latent states for the most informed transition prediction, whereas closed-loop prior rollouts evaluate predictive behavior during imagination. Actions are selected greedily, while dynamics predictions remain stochastic. We set the number of rollouts to $R=10$.

\paragraph{Computation of Proprioceptive Discrepancy.} Proprioceptive discrepancy is measured as the mean $\ell_1$ distance between predicted and ground-truth body positions, where ground truth is obtained by simulating the predicted action sequence from a given proprioceptive start state.

\begin{figure}[htb!]
\centering
\input{figs/phys_diff_plots/phys_diff.tex}
\caption{Proprioceptive discrepancy versus prior uncertainty for ID and OOD initial states in RSSM and Cat-RSSM. Solid lines show means over $R \cdot K \cdot S = 500$ samples per time step and setting; shaded areas denote standard deviation. Although uncertainty captures initial distribution shifts, it does not track the compounding proprioceptive errors that emerge during latent rollouts.}
\label{fig:phys_diff}
\end{figure}

We report the stepwise mean and standard deviation of proprioceptive discrepancy and prior uncertainty (Fig.~\ref{fig:phys_diff}) and discuss RSSM and Cat-RSSM jointly due to their similar behavior. Posterior-informed rollouts remain accurate for ID initial states, while initially high OOD discrepancy decreases as observations refine the latent state. Closed-loop prior rollouts accumulate proprioceptive discrepancy in both settings, but uncertainty does not reflect this error growth: ID uncertainty remains low and stable, whereas OOD uncertainty starts elevated, but rapidly converges to ID levels. This behavior can be attributed to the attractor effect observed for OOD rollouts in Section~\ref{sec:attr_eval}. While ensemble-based uncertainty appears locally informative, RSSM latent dynamics favor regions that are well-supported by the training distribution and exhibit low ensemble disagreement during prolonged rollouts, preventing uncertainty from reliably tracking compounding model error.

%-------------------------------------------------------------

\subsection{Reward Evaluation}
\label{sec:reward_eval}

To address \textit{Question}~\ref{prob3}, we evaluate how latent-environment misalignment affects reward prediction. We compute a stepwise signed reward discrepancy $r^{\text{pred}}_t - r^{\text{sim}}_t$ over the collected closed-loop prior and posterior-informed rollouts, where $r^{\text{pred}}_t$ denotes the model-predicted reward along the latent rollout and $r^{\text{sim}}_t$ the reward obtained by executing the same action sequence in the environment.

We report the stepwise mean and standard deviation of signed reward discrepancy (Fig.~\ref{fig:rewards}) and discuss RSSM and Cat-RSSM jointly due to their similar behavior. Posterior-informed rollouts accurately predict rewards under observation updates, whereas closed-loop prior rollouts systematically overestimate them. Together with the attractor behavior observed in Section~\ref{sec:attr_eval}, this suggests that latent rollouts are biased toward well-supported regions that appear to coincide with high-performing behaviors in dense-reward tasks, contributing to overly optimistic reward predictions.

\begin{figure}[htb!]
\centering
\input{figs/reward_plots/reward_diff.tex}
\caption{Signed reward discrepancy in RSSM and Cat-RSSM. Solid lines show means over $N \cdot S = 5000$ samples per time step and setting; shaded areas denote standard deviation. Closed-loop prior rollouts systematically overestimate rewards, consistent with an attractor bias toward well-supported latent regions that appear to coincide with high-performing behaviors.}
\label{fig:rewards}
\end{figure}
%%%%%%%%%%%%%%%%%%%%%%%%%%%%%%%%%%%%%%%%%%%%%%%%%%%%%%%%%%%%%%%%
%% Section: Conclusion
%%%%%%%%%%%%%%%%%%%%%%%%%%%%%%%%%%%%%%%%%%%%%%%%%%%%%%%%%%%%%%%%
\section{Conclusion}
\label{sec:conclusion}

Our findings reveal a structural limitation of epistemic uncertainty estimation in RSSMs: While ensemble disagreement has been shown to capture \textit{local} transition reliability~\citep{lakshminarayanan2017simple,chua2018deep}, this relationship does not necessarily extend to \textit{global} trajectory-level model error during prolonged latent rollouts. We provide evidence for an attractor behavior in RSSM latent dynamics, where rollouts are biased toward latent regions well-supported by the learned dynamics, as indicated by low ensemble disagreement. As a result, epistemic uncertainty can diminish despite increasing discrepancies from true environment dynamics, masking compounding model error. In dense-reward tasks, this behavior may contribute to overly optimistic reward predictions, as attractor regions appear to align with high-reward behaviors.

\paragraph{Future Work.} We hypothesize that this limitation originates from inductive biases in VAE-based latent dynamics and cannot be resolved through improved uncertainty estimation alone. Instead, addressing it may require changes to latent dynamics modeling, \eg through latent space restructuring~\citep{li2024fld,becker2023coral}, more principled variational inference~\citep{becker2022vrkn}, or alternative latent parametrizations. More broadly, our findings motivate further investigation of epistemic uncertainty in latent dynamics models beyond RSSMs. A formal characterization of the attractor mechanism remains an important direction for future work.

\section*{Acknowledgments}

We thank Tim Elsner and Leif Kobbelt for their support during an earlier research project that inspired this paper. This work was partially supported by the Robotics Institute Germany (RIG). The authors gratefully acknowledge the computing time provided to them at the NHR Center NHR4CES at RWTH Aachen University (project number p0022301). This is funded by the Federal Ministry of Research, Technology and Space, and the state governments participating on the basis of the resolutions of the GWK for national high performance computing at universities (\url{www.nhr-verein.de/unsere-partner}).

%%%%%%%%%%%%%%%%%%%%%%%%%%%%%%%%%%%%%%%%%%%%%%%%%%%%%%%%%%%%%%%%
%% NOTE: THIS MARKS THE END OF THE "MAIN TEXT"
%%%%%%%%%%%%%%%%%%%%%%%%%%%%%%%%%%%%%%%%%%%%%%%%%%%%%%%%%%%%%%%%

%%%%%%%%%%%%%%%%%%%%%%%%%%%%%%%%%%%%%%%%%%%%%%%%%%%%%%%%%%%%%%%%
%% Bibliography
%%%%%%%%%%%%%%%%%%%%%%%%%%%%%%%%%%%%%%%%%%%%%%%%%%%%%%%%%%%%%%%%
\bibliography{main}

\begin{thebibliography}{53}
\providecommand{\natexlab}[1]{#1}
\providecommand{\url}[1]{\texttt{#1}}
\expandafter\ifx\csname urlstyle\endcsname\relax
  \providecommand{\doi}[1]{DOI: #1}\else
  \providecommand{\doi}{DOI: \begingroup \urlstyle{rm}\Url}\fi

\bibitem[Assran et~al.(2025)Assran, Bardes, Fan, Garrido, Howes, Muckley, Rizvi, Roberts, Sinha, Zholus, et~al.]{assran2025v}
Mido Assran, Adrien Bardes, David Fan, Quentin Garrido, Russell Howes, Matthew Muckley, Ammar Rizvi, Claire Roberts, Koustuv Sinha, Artem Zholus, et~al.
\newblock V-jepa 2: Self-supervised video models enable understanding, prediction and planning.
\newblock \emph{arXiv preprint arXiv:2506.09985}, 2025.

\bibitem[Barto \& Sutton(2018)Barto and Sutton]{andrew2018reinforcement}
Andrew Barto and Richard Sutton.
\newblock \emph{Reinforcement learning: an introduction}.
\newblock The MIT Press, 2018.

\bibitem[Becker \& Neumann(2022)Becker and Neumann]{becker2022vrkn}
Philipp Becker and Gerhard Neumann.
\newblock On uncertainty in deep state space models for model-based reinforcement learning.
\newblock \emph{Transactions on Machine Learning Research (TMLR)}, 2022.

\bibitem[Becker et~al.(2024)Becker, Mossburger, Otto, and Neumann]{becker2023coral}
Philipp Becker, Sebastian Mossburger, Fabian Otto, and Gerhard Neumann.
\newblock Combining reconstruction and contrastive methods for multimodal representations in rl.
\newblock \emph{Reinforcement Learning Conference (RLC)}, 2024.

\bibitem[Berner et~al.(2019)Berner, Brockman, Chan, Cheung, D{\k{e}}biak, Dennison, Farhi, Fischer, Hashme, Hesse, et~al.]{berner2019dota}
Christopher Berner, Greg Brockman, Brooke Chan, Vicki Cheung, Przemys{\l}aw D{\k{e}}biak, Christy Dennison, David Farhi, Quirin Fischer, Shariq Hashme, Chris Hesse, et~al.
\newblock Dota 2 with large scale deep reinforcement learning.
\newblock \emph{arXiv preprint arXiv:1912.06680}, 2019.

\bibitem[Buckman et~al.(2018)Buckman, Hafner, Tucker, Brevdo, and Lee]{buckman2018sample}
Jacob Buckman, Danijar Hafner, George Tucker, Eugene Brevdo, and Honglak Lee.
\newblock Sample-efficient reinforcement learning with stochastic ensemble value expansion.
\newblock \emph{Advances in Neural Information Processing Systems (NeurIPS)}, 31, 2018.

\bibitem[Chen et~al.(2022)Chen, Wu, Yoon, and Ahn]{chen2022transdreamer}
Chang Chen, Yi-Fu Wu, Jaesik Yoon, and Sungjin Ahn.
\newblock Transdreamer: Reinforcement learning with transformer world models.
\newblock \emph{arXiv preprint arXiv:2202.09481}, 2022.

\bibitem[Chua et~al.(2018)Chua, Calandra, McAllister, and Levine]{chua2018deep}
Kurtland Chua, Roberto Calandra, Rowan McAllister, and Sergey Levine.
\newblock Deep reinforcement learning in a handful of trials using probabilistic dynamics models.
\newblock \emph{Advances in Neural Information Processing Systems (NeurIPS)}, 31, 2018.

\bibitem[Degrave et~al.(2022)Degrave, Felici, Buchli, Neunert, Tracey, Carpanese, Ewalds, Hafner, Abdolmaleki, de~Las~Casas, et~al.]{degrave2022magnetic}
Jonas Degrave, Federico Felici, Jonas Buchli, Michael Neunert, Brendan Tracey, Francesco Carpanese, Timo Ewalds, Roland Hafner, Abbas Abdolmaleki, Diego de~Las~Casas, et~al.
\newblock Magnetic control of tokamak plasmas through deep reinforcement learning.
\newblock \emph{Nature}, 602\penalty0 (7897):\penalty0 414--419, 2022.

\bibitem[Deisenroth \& Rasmussen(2011)Deisenroth and Rasmussen]{deisenroth2011pilco}
Marc Deisenroth and Carl~E Rasmussen.
\newblock Pilco: A model-based and data-efficient approach to policy search.
\newblock In \emph{Proceedings of the 28th International Conference on Machine Learning (ICML-11)}, pp.\  465--472, 2011.

\bibitem[Filos et~al.(2022)Filos, V{\'e}rtes, Marinho, Farquhar, Borsa, Friesen, Behbahani, Schaul, Barreto, and Osindero]{filos2022modelvalueinconsistency}
Angelos Filos, Eszter V{\'e}rtes, Zita Marinho, Gregory Farquhar, Diana Borsa, Abram Friesen, Feryal Behbahani, Tom Schaul, Andr{\'e} Barreto, and Simon Osindero.
\newblock Model-value inconsistency as a signal for epistemic uncertainty.
\newblock \emph{International Conference on Learning Representations (ICLR)}, 2022.

\bibitem[Finn \& Levine(2017)Finn and Levine]{finn2017deep}
Chelsea Finn and Sergey Levine.
\newblock Deep visual foresight for planning robot motion.
\newblock In \emph{2017 IEEE international conference on robotics and automation (ICRA)}, pp.\  2786--2793. IEEE, 2017.

\bibitem[Frauenknecht et~al.(2024)Frauenknecht, Eisele, Subhasish, Solowjow, and Trimpe]{frauenknecht2024macura}
Bernd Frauenknecht, Artur Eisele, Devdutt Subhasish, Friedrich Solowjow, and Sebastian Trimpe.
\newblock Trust the model where it trusts itself--model-based actor-critic with uncertainty-aware rollout adaption.
\newblock \emph{International Conference on Machine Learning (ICML)}, 2024.

\bibitem[Frauenknecht et~al.(2025)Frauenknecht, Subhasish, Solowjow, and Trimpe]{frauenknecht2025infoprop}
Bernd Frauenknecht, Devdutt Subhasish, Friedrich Solowjow, and Sebastian Trimpe.
\newblock On rollouts in model-based reinforcement learning.
\newblock \emph{International Conference on Learning Representations (ICLR)}, 2025.

\bibitem[Frauenknecht et~al.(2026)Frauenknecht, Kesper, Mayfrank, Hose, and Trimpe]{frauenknecht2026uncertainty}
Bernd Frauenknecht, Lukas Kesper, Daniel Mayfrank, Henrik Hose, and Sebastian Trimpe.
\newblock Uncertainty-aware predictive safety filters for probabilistic neural network dynamics.
\newblock \emph{Reinforcement Learning Conference}, 2026.

\bibitem[Guo et~al.(2022)Guo, Thakoor, P{\^\i}slar, Avila~Pires, Altch{\'e}, Tallec, Saade, Calandriello, Grill, Tang, et~al.]{guo2022byol}
Zhaohan Guo, Shantanu Thakoor, Miruna P{\^\i}slar, Bernardo Avila~Pires, Florent Altch{\'e}, Corentin Tallec, Alaa Saade, Daniele Calandriello, Jean-Bastien Grill, Yunhao Tang, et~al.
\newblock Byol-explore: Exploration by bootstrapped prediction.
\newblock \emph{Advances in neural information processing systems}, 35:\penalty0 31855--31870, 2022.

\bibitem[Gupta et~al.(2021)Gupta, Sharma, Jain, Liang, Broeck, and Singla]{gupta2021towards}
Rushil Gupta, Vishal Sharma, Yash Jain, Yitao Liang, Guy Van~den Broeck, and Parag Singla.
\newblock Towards an interpretable latent space in structured models for video prediction.
\newblock \emph{arXiv preprint arXiv:2107.07713}, 2021.

\bibitem[Ha \& Schmidhuber(2018)Ha and Schmidhuber]{ha2018world}
David Ha and J{\"u}rgen Schmidhuber.
\newblock World models.
\newblock \emph{Conference on Neural Information Processing Systems (NeurIPS)}, 2\penalty0 (3):\penalty0 440, 2018.

\bibitem[Hafner et~al.(2019{\natexlab{a}})Hafner, Lillicrap, Ba, and Norouzi]{hafner2019dreamerv1}
Danijar Hafner, Timothy Lillicrap, Jimmy Ba, and Mohammad Norouzi.
\newblock Dream to control: Learning behaviors by latent imagination.
\newblock \emph{arXiv preprint arXiv:1912.01603}, 2019{\natexlab{a}}.

\bibitem[Hafner et~al.(2019{\natexlab{b}})Hafner, Lillicrap, Fischer, Villegas, Ha, Lee, and Davidson]{hafner2019planet}
Danijar Hafner, Timothy Lillicrap, Ian Fischer, Ruben Villegas, David Ha, Honglak Lee, and James Davidson.
\newblock Learning latent dynamics for planning from pixels.
\newblock In \emph{International Conference on Machine Learning (ICML)}, pp.\  2555--2565. PMLR, 2019{\natexlab{b}}.

\bibitem[Hafner et~al.(2020)Hafner, Lillicrap, Norouzi, and Ba]{hafner2020dreamerv2}
Danijar Hafner, Timothy Lillicrap, Mohammad Norouzi, and Jimmy Ba.
\newblock Mastering atari with discrete world models.
\newblock \emph{arXiv preprint arXiv:2010.02193}, 2020.

\bibitem[Hafner et~al.(2023)Hafner, Pasukonis, Ba, and Lillicrap]{hafner2023dreamerv3}
Danijar Hafner, Jurgis Pasukonis, Jimmy Ba, and Timothy Lillicrap.
\newblock Mastering diverse domains through world models.
\newblock \emph{arXiv preprint arXiv:2301.04104}, 2023.

\bibitem[Hansen et~al.(2022)Hansen, Wang, and Su]{hansen2022tdmpc}
Nicklas Hansen, Xiaolong Wang, and Hao Su.
\newblock Temporal difference learning for model predictive control.
\newblock \emph{International Conference on Machine Learning (ICML)}, 2022.

\bibitem[Hansen et~al.(2024)Hansen, Su, and Wang]{hansen2023tdmpc2}
Nicklas Hansen, Hao Su, and Xiaolong Wang.
\newblock Td-mpc2: Scalable, robust world models for continuous control.
\newblock \emph{International Conference on Learning Representations (ICLR)}, 2024.

\bibitem[Janner et~al.(2019)Janner, Fu, Zhang, and Levine]{janner2019mbpo}
Michael Janner, Justin Fu, Marvin Zhang, and Sergey Levine.
\newblock When to trust your model: Model-based policy optimization.
\newblock \emph{Advances in Neural Information Processing Systems (NeurIPS)}, 32, 2019.

\bibitem[Jarrett et~al.(2022)Jarrett, Tallec, Altch{\'e}, Mesnard, Munos, and Valko]{jarrett2022curiosity}
Daniel Jarrett, Corentin Tallec, Florent Altch{\'e}, Thomas Mesnard, R{\'e}mi Munos, and Michal Valko.
\newblock Curiosity in hindsight: Intrinsic exploration in stochastic environments.
\newblock \emph{arXiv preprint arXiv:2211.10515}, 2022.

\bibitem[Lakshminarayanan et~al.(2017)Lakshminarayanan, Pritzel, and Blundell]{lakshminarayanan2017simple}
Balaji Lakshminarayanan, Alexander Pritzel, and Charles Blundell.
\newblock Simple and scalable predictive uncertainty estimation using deep ensembles.
\newblock \emph{Advances in Neural Information Processing Systems (NeurIPS)}, 30, 2017.

\bibitem[Levine \& Koltun(2013)Levine and Koltun]{levine2013guided}
Sergey Levine and Vladlen Koltun.
\newblock Guided policy search.
\newblock In \emph{International Conference on Machine Learning (ICML)}, pp.\  1--9. PMLR, 2013.

\bibitem[Li et~al.(2024)Li, Stanger-Jones, Heim, and Kim]{li2024fld}
Chenhao Li, Elijah Stanger-Jones, Steve Heim, and Sangbae Kim.
\newblock Fld: Fourier latent dynamics for structured motion representation and learning.
\newblock \emph{International Conference on Learning Representations (ICLR)}, 2024.

\bibitem[Liu et~al.(2026)Liu, Peng, Huang, and Tian]{liu2026perceiving}
Zhenxian Liu, Peixi Peng, Yangru Huang, and Yonghong Tian.
\newblock Perceiving the knowledge boundary: Uncertainty-guided exploration and imagination for world models.
\newblock In \emph{Proceedings of the AAAI Conference on Artificial Intelligence}, volume~40, 2026.

\bibitem[Micheli et~al.(2022)Micheli, Alonso, and Fleuret]{micheli2022transformers}
Vincent Micheli, Eloi Alonso, and Fran{\c{c}}ois Fleuret.
\newblock Transformers are sample-efficient world models.
\newblock \emph{arXiv preprint arXiv:2209.00588}, 2022.

\bibitem[Nagabandi et~al.(2018)Nagabandi, Kahn, Fearing, and Levine]{nagabandi2018neural}
Anusha Nagabandi, Gregory Kahn, Ronald~S Fearing, and Sergey Levine.
\newblock Neural network dynamics for model-based deep reinforcement learning with model-free fine-tuning.
\newblock In \emph{2018 IEEE International Conference on Robotics and Automation (ICRA)}, pp.\  7559--7566. IEEE, 2018.

\bibitem[Nakamura et~al.(2025)Nakamura, Peters, and Bajcsy]{nakamura2025generalizing}
Kensuke Nakamura, Lasse Peters, and Andrea Bajcsy.
\newblock Generalizing safety beyond collision-avoidance via latent-space reachability analysis.
\newblock \emph{arXiv preprint arXiv:2502.00935}, 2025.

\bibitem[Nielsen(2019)]{nielsen2019jensen}
Frank Nielsen.
\newblock On the jensen--shannon symmetrization of distances relying on abstract means.
\newblock \emph{Entropy}, 21\penalty0 (5):\penalty0 485, 2019.

\bibitem[Pathak et~al.(2017)Pathak, Agrawal, Efros, and Darrell]{pathak2017curiosity}
Deepak Pathak, Pulkit Agrawal, Alexei~A Efros, and Trevor Darrell.
\newblock Curiosity-driven exploration by self-supervised prediction.
\newblock In \emph{International conference on machine learning}, pp.\  2778--2787. PMLR, 2017.

\bibitem[Peper et~al.(2025)Peper, Mao, Geng, Pan, and Ruchkin]{peper2025four}
Jordan Peper, Zhenjiang Mao, Yuang Geng, Siyuan Pan, and Ivan Ruchkin.
\newblock Four principles for physically interpretable world models.
\newblock \emph{IEEE International Conference on Robotics and Automation (ICRA)}, 2025.

\bibitem[Sancaktar et~al.(2022)Sancaktar, Blaes, and Martius]{sancaktar2022curious}
Cansu Sancaktar, Sebastian Blaes, and Georg Martius.
\newblock Curious exploration via structured world models yields zero-shot object manipulation.
\newblock \emph{Advances in Neural Information Processing Systems (NeurIPS)}, 35:\penalty0 24170--24183, 2022.

\bibitem[Sancaktar et~al.(2025)Sancaktar, Gumbsch, Zadaianchuk, Kolev, and Martius]{sancaktar2025sensei}
Cansu Sancaktar, Christian Gumbsch, Andrii Zadaianchuk, Pavel Kolev, and Georg Martius.
\newblock Sensei: Semantic exploration guided by foundation models to learn versatile world models.
\newblock \emph{arXiv preprint arXiv:2503.01584}, 2025.

\bibitem[Schwarzer et~al.(2020)Schwarzer, Anand, Goel, Hjelm, Courville, and Bachman]{schwarzer2020data}
Max Schwarzer, Ankesh Anand, Rishab Goel, R~Devon Hjelm, Aaron Courville, and Philip Bachman.
\newblock Data-efficient reinforcement learning with self-predictive representations.
\newblock \emph{arXiv preprint arXiv:2007.05929}, 2020.

\bibitem[Sekar et~al.(2020)Sekar, Rybkin, Daniilidis, Abbeel, Hafner, and Pathak]{sekar2020plan2explore}
Ramanan Sekar, Oleh Rybkin, Kostas Daniilidis, Pieter Abbeel, Danijar Hafner, and Deepak Pathak.
\newblock Planning to explore via self-supervised world models.
\newblock In \emph{International Conference on Machine Learning (ICML)}, pp.\  8583--8592. PMLR, 2020.

\bibitem[Seo et~al.(2025)Seo, Nakamura, and Bajcsy]{seo2025unisafe}
Junwon Seo, Kensuke Nakamura, and Andrea Bajcsy.
\newblock Uncertainty-aware latent safety filters for avoiding out-of-distribution failures.
\newblock \emph{Conference on Robot Learning (CoRL)}, 2025.

\bibitem[Seyde et~al.(2020)Seyde, Schwarting, Karaman, and Rus]{seyde2020love}
Tim Seyde, Wilko Schwarting, Sertac Karaman, and Daniela Rus.
\newblock Learning to plan optimistically: Uncertainty-guided deep exploration via latent model ensembles.
\newblock \emph{Computing Research Repository (CoRR)}, 2020.

\bibitem[Sukhija et~al.(2023)Sukhija, Treven, Sancaktar, Blaes, Coros, and Krause]{NEURIPS2023_77b5aaf2}
Bhavya Sukhija, Lenart Treven, Cansu Sancaktar, Sebastian Blaes, Stelian Coros, and Andreas Krause.
\newblock Optimistic active exploration of dynamical systems.
\newblock In A.~Oh, T.~Naumann, A.~Globerson, K.~Saenko, M.~Hardt, and S.~Levine (eds.), \emph{Advances in Neural Information Processing Systems}, volume~36, pp.\  38122--38153. Curran Associates, Inc., 2023.

\bibitem[Tassa et~al.(2018)Tassa, Doron, Muldal, Erez, Li, de~Las~Casas, Budden, Abdolmaleki, Merel, Lefrancq, Lillicrap, and Riedmiller]{tassa2018dmc}
Yuval Tassa, Yotam Doron, Alistair Muldal, Tom Erez, Yazhe Li, Diego de~Las~Casas, David Budden, Abbas Abdolmaleki, Josh Merel, Andrew Lefrancq, Timothy Lillicrap, and Martin Riedmiller.
\newblock Deepmind control suite, 2018.

\bibitem[Todorov et~al.(2012)Todorov, Erez, and Tassa]{todorov2012mujoco}
Emanuel Todorov, Tom Erez, and Yuval Tassa.
\newblock Mujoco: A physics engine for model-based control.
\newblock In \emph{2012 IEEE/RSJ International Conference on Intelligent Robots and Systems}, pp.\  5026--5033. IEEE, 2012.

\bibitem[Vignola et~al.(2026)Vignola, Lee, Prajapat, Wendl, Zeilinger, Krause, and As]{vignola2026sampling}
Luca Vignola, Bruce~D Lee, Manish Prajapat, Manuel Wendl, Melanie Zeilinger, Andreas Krause, and Yarden As.
\newblock Sampling-based safe reinforcement learning.
\newblock \emph{ICML Workshop on Decision-Making from Offline Datasets to Online Adaptation}, 2026.

\bibitem[Wang et~al.(2020)Wang, Zhang, Zhao, Zhao, and Hao]{wang2020dmve}
Junjie Wang, Qichao Zhang, Dongbin Zhao, Mengchen Zhao, and Jianye Hao.
\newblock Dynamic horizon value estimation for model-based reinforcement learning.
\newblock \emph{arXiv preprint arXiv:2009.09593}, 2020.

\bibitem[Williams et~al.(2017)Williams, Wagener, Goldfain, Drews, Rehg, Boots, and Theodorou]{williams2017information}
Grady Williams, Nolan Wagener, Brian Goldfain, Paul Drews, James~M Rehg, Byron Boots, and Evangelos~A Theodorou.
\newblock Information theoretic mpc for model-based reinforcement learning.
\newblock In \emph{2017 IEEE International Conference on Robotics and Automation (ICRA)}, pp.\  1714--1721. IEEE, 2017.

\bibitem[Yu et~al.(2023)Yu, Zou, Yang, Ma, Li, Yin, Chen, and Duan]{yu2023safe}
Dongjie Yu, Wenjun Zou, Yujie Yang, Haitong Ma, Shengbo~Eben Li, Yuming Yin, Jianyu Chen, and Jingliang Duan.
\newblock Safe model-based reinforcement learning with an uncertainty-aware reachability certificate.
\newblock \emph{IEEE Transactions on Automation Science and Engineering}, 21\penalty0 (3):\penalty0 4129--4142, 2023.

\bibitem[Yu et~al.(2020)Yu, Thomas, Yu, Ermon, Zou, Levine, Finn, and Ma]{yu2020mopo}
Tianhe Yu, Garrett Thomas, Lantao Yu, Stefano Ermon, James~Y Zou, Sergey Levine, Chelsea Finn, and Tengyu Ma.
\newblock Mopo: Model-based offline policy optimization.
\newblock \emph{Advances in Neural Information Processing Systems (NeurIPS)}, 33:\penalty0 14129--14142, 2020.

\bibitem[Zhang et~al.(2019)Zhang, Vikram, Smith, Abbeel, Johnson, and Levine]{zhang2019solar}
Marvin Zhang, Sharad Vikram, Laura Smith, Pieter Abbeel, Matthew Johnson, and Sergey Levine.
\newblock Solar: Deep structured representations for model-based reinforcement learning.
\newblock In \emph{International conference on machine learning}, pp.\  7444--7453. PMLR, 2019.

\bibitem[Zhou et~al.(2024)Zhou, Pan, LeCun, and Pinto]{zhou2024dino}
Gaoyue Zhou, Hengkai Pan, Yann LeCun, and Lerrel Pinto.
\newblock Dino-wm: World models on pre-trained visual features enable zero-shot planning.
\newblock \emph{International Conference on Machine Learning (ICML)}, 2024.

\bibitem[Zhu et~al.(2020)Zhu, Zhang, Lee, and Zhang]{zhu2020bird}
Guangxiang Zhu, Minghao Zhang, Honglak Lee, and Chongjie Zhang.
\newblock Bridging imagination and reality for model-based deep reinforcement learning.
\newblock \emph{Advances in Neural Information Processing Systems (NeurIPS)}, 33:\penalty0 8993--9006, 2020.

\end{thebibliography}
\bibliographystyle{rlj}

%%%%%%%%%%%%%%%%%%%%%%%%%%%%%%%%%%%%%%%%%%%%%%%%%%%%%%%%%%%%%%%%
%% Supplementary Materials
%%%%%%%%%%%%%%%%%%%%%%%%%%%%%%%%%%%%%%%%%%%%%%%%%%%%%%%%%%%%%%%%
\beginSupplementaryMaterials
%%%%%%%%%%%%%%%%%%%%%%%%%%%%%%%%%%%%%%%%%%%%%%%%%%%%%%%%%%%%%%%%
%% Supplementary
%%%%%%%%%%%%%%%%%%%%%%%%%%%%%%%%%%%%%%%%%%%%%%%%%%%%%%%%%%%%%%%%

\section{Experiment Setup}
\label{sec:supp_exp}

Our evaluation code\footnote{\url{https://github.com/jberger999/biased-dreams}} follows~\citet{becker2022vrkn, becker2023coral}, which in turn builds on~\citet{hafner2019dreamerv1, hafner2020dreamerv2}. For details on Infoprop, see~\cite{frauenknecht2025infoprop}.

% --------------------------------------------------------------
\subsection{Training Details}

\paragraph{Environments.} All environments use an action repeat of 2. Images are $64\times64$ pixels with 5-bit color depth, following~\cite{hafner2019planet}.

\paragraph{Architecture.} For RSSM, we use a stochastic state size of 30 and a deterministic state size for 200. For Cat-RSSM, the stochastic state is replaced by 32 categoricals with 32 classes each. Independent of architecture, all layers have 300 units, with standard encoders and decoders using ReLU. Note that Cat-RSSM, as proposed by \citet{hafner2020dreamerv2}, also increases model size, which we omit to ensure a fair comparison with RSSM. The transition matrix output is transformed via a sigmoid-based function, as described in \cite{becker2022vrkn}.

\paragraph{Proprioceptive Decoder and Ensemble.} The proprioceptive state decoder follows the decoder architecture in \cite{becker2022vrkn}, with 3 layers of size 300 and ELU activation. Each angle $\theta$ in the proprioceptive state is encoded as $(\sin\theta, \cos\theta)$ and reconstructed for evaluation and simulation from $\text{arctan2}(y_1,y_2)$ given network angle predictions $(y_1,y_2)$. Each member of the ensemble has 5 linear layers of size 300, intermediately normalized with LayerNorm, and activated with ELU. The ensemble predicts the next deterministic state, following standard ensemble-based implementations~\citep{seo2025unisafe}.

\paragraph{Loss.} RSSM and actor-critic losses follow \citet{hafner2019dreamerv1}, including a separate reconstruction loss for the proprioceptive decoder, when applicable. The ensemble loss is the negative Gaussian log-likelihood of the target state. During ensemble loss computation, inputs and targets are detached to prevent gradient flow through the model and actor. For Cat-RSSM, we additionally employ KL balancing as proposed by \cite{hafner2020dreamerv2}, separately scaling posterior and prior entropy in the KL term with $\alpha = 0.8$. We omit adding action entropy in the policy loss to focus on architectural differences between RSSM and Cat-RSSM, rather than differences in the learning objective.

\paragraph{Training.} We begin training with 5 random episodes and collect one additional episode every 100 model update steps using exploration noise of 0.3. Each update samples 50 subsequences of length 50 uniformly from all collected data. Adam is used with learning rates and gradient clipping as in \citet{hafner2019dreamerv1}: latent dynamics, $6 \cdot 10^{-4}$; actor-critic, $8 \cdot 10^{-5}$; gradient norm clipped at 100. The ensemble uses a separate optimizer with the same optimizer hyperparameters as the latent dynamics model.

% --------------------------------------------------------------
\subsection{Evaluation Details}

\paragraph{Aggregation.} For each DMC Suite task, we perform $S=5$ training runs with different random seeds to account for training variability while keeping computational cost manageable. For the proprioceptive and reward discrepancy plots (Figs.~\ref{fig:phys_diff} and~\ref{fig:rewards}), we report the stepwise mean and standard deviation over all collected samples. As generic descriptive statistics, mean and standard deviation summarize the overall tendency and variability of rollout distributions without imposing a specific distributional model.

\paragraph{Rollout Types.} Posterior-informed rollouts are obtained by applying a one-step prior transition (Eq.~\ref{eq:post_inf_rollouts}) from each posterior state inferred by the representation model (Eq.~\ref{eq:repr_model}). Compared to posterior rollouts, this reduces observation-induced noise in proprioceptive reconstructions, enabling a cleaner assessment of predictive dynamics accuracy. Closed-loop prior rollouts start from a posterior state and evolve under a reactive greedy policy with stochastic latent dynamics. Open-loop prior rollouts instead replay the action sequence of a posterior rollout under stochastic latent dynamics, isolating the effects of model stochasticity and error accumulation. In practice, these rollouts tend to induce state-action mismatches, making them well-suited for identifying highly uncertain transitions.

% --------------------------------------------------------------
\section{Additional Results}

% --------------------------------------------------------------
\subsection{Attractor Evaluation}
\label{sec:supp_attractor_eval}

The remaining DMC Suite environments exhibit similar latent rollout patterns in the embedded PCA space consistent with those observed in the Cheetah Run evaluation presented in Section~\ref{sec:attr_eval}, across both RSSM and Cat-RSSM (Fig.~\ref{fig:attractor_rest}). These observations indicate that the observed attractor behavior is not specific to a single task or architecture, but appears consistently across all evaluated settings.

\begingroup
\setlength{\tabcolsep}{1pt}
\begin{figure}[htb!]
    \centering
    \begin{tabular}{cc|c}
        & RSSM & Cat-RSSM\\

        &
        \hspace{1.2cm} \makebox[0pt][c]{\footnotesize ID} \hspace{2.2cm} \makebox[0pt][c]{\footnotesize OOD} \hspace{1.5cm} &
        \hspace{1.2cm} \makebox[0pt][c]{\footnotesize ID} \hspace{2.2cm} \makebox[0pt][c]{\footnotesize OOD} \hspace{1.5cm} \\

        \raisebox{0.5\height}{\rotatebox{90}{\footnotesize Cheetah Run}} &
        \includegraphics[width=0.48\linewidth]{figs/pca_plots/rssm/pca-rssm-cheetah_run.pdf} &
        \includegraphics[width=0.48\linewidth]{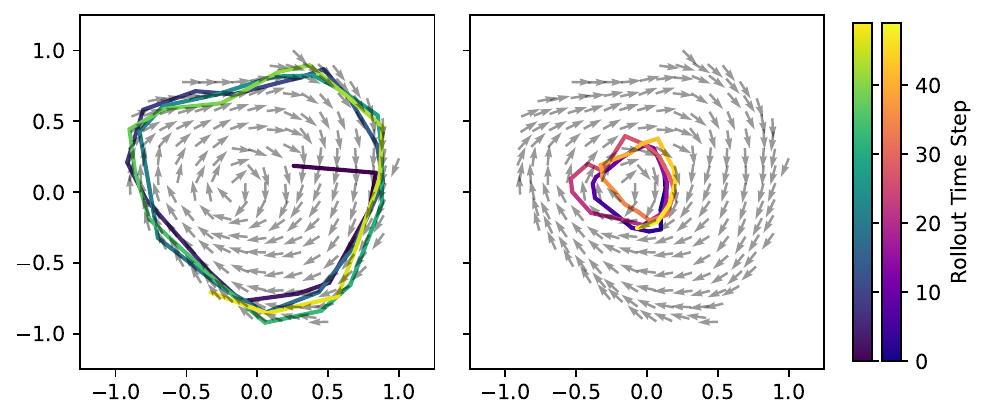}\\

        \raisebox{0.2\height}{\rotatebox{90}{\footnotesize Cartpole Swingup}} &
        \includegraphics[width=0.48\linewidth]{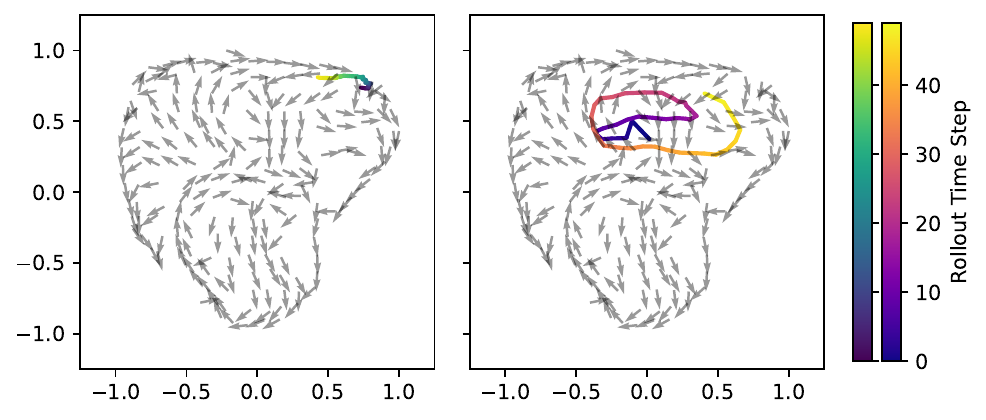} &
        \includegraphics[width=0.48\linewidth]{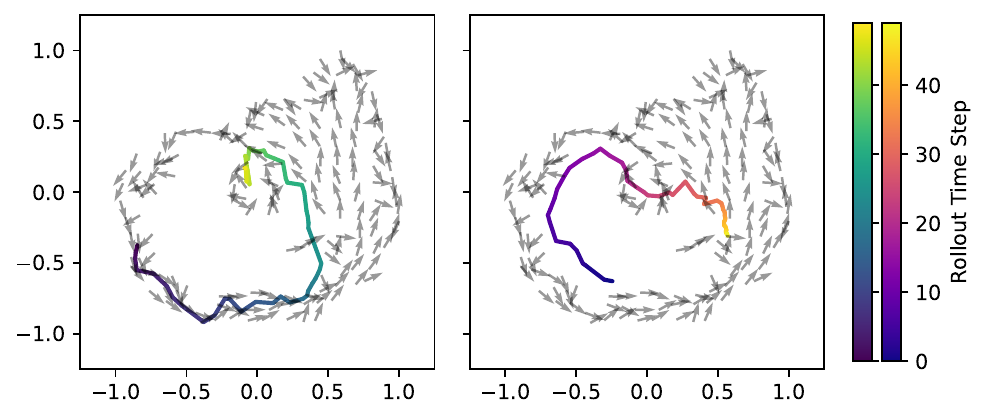}\\
        
        \raisebox{0.5\height}{\rotatebox{90}{\footnotesize Hopper Hop}} &
        \includegraphics[width=0.48\linewidth]{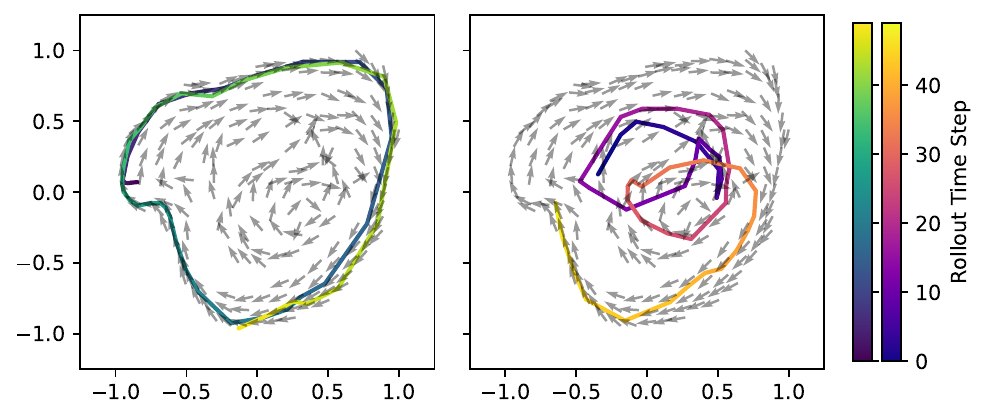} &
        \includegraphics[width=0.48\linewidth]{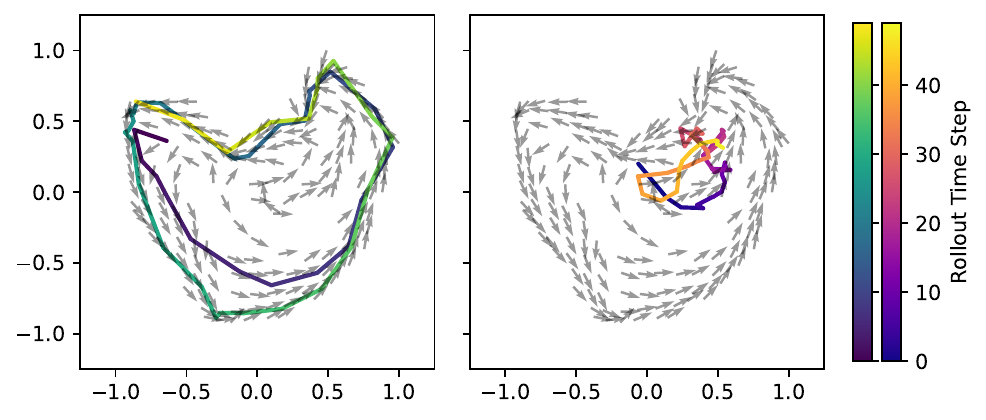}\\
        
        \raisebox{0.5\height}{\rotatebox{90}{\footnotesize Walker Run}} &
        \includegraphics[width=0.48\linewidth]{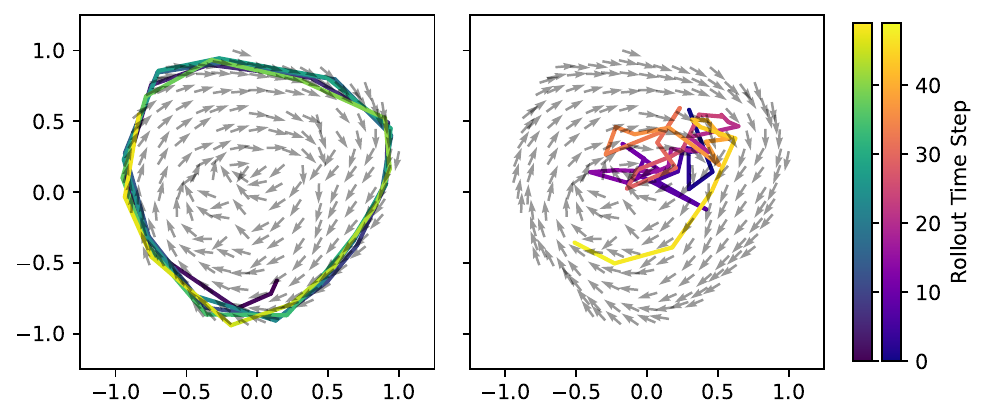} &
        \includegraphics[width=0.48\linewidth]{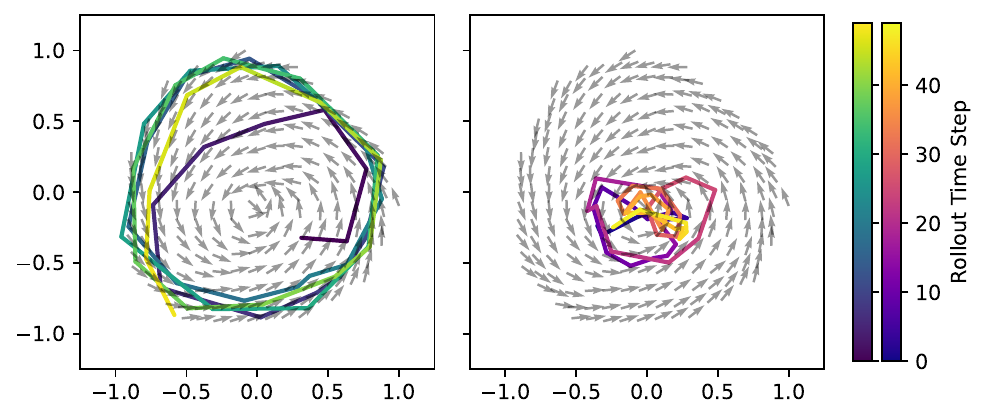}\\
    \end{tabular}
    \caption{Attractor analysis of RSSM and Cat-RSSM across DMC Suite environments.}
    \label{fig:attractor_rest}
\end{figure}
\endgroup

% --------------------------------------------------------------
\newpage
\subsection{Proprioceptive Discrepancy}

We report Dreamer's RL performance using RSSM and Cat-RSSM, with and without a PD, separately for each task (Fig.~\ref{fig:rl_performance_separated}). The per-task results are consistent with the aggregated plots in Section~\ref{sec:phys_discr}, showing that the incorporating the PD into the latent dynamics model has negligible impact on Dreamer's performance.

\begin{figure}[htb!]
\centering
\input{figs/rl_performance_plots/rl_performance.tex}
{\small Environment Step ($\times 10^6$)}
\caption{Dreamer's performance is largely unchanged by the use of the proprioceptive state decoder (PD) between approaches. Results are averaged over 5 seeds; shaded areas show 95\% confidence intervals.}
\label{fig:rl_performance_separated}
\end{figure}

% --------------------------------------------------------------
\subsection{Image and Proprioceptive Reconstructions}

\begin{figure}[htb!]
\centering
\begin{tikzpicture}
% ----- First row -----
\node[anchor=south west, inner sep=0] (img1) at (0,0)
    {\includegraphics[width=13cm]{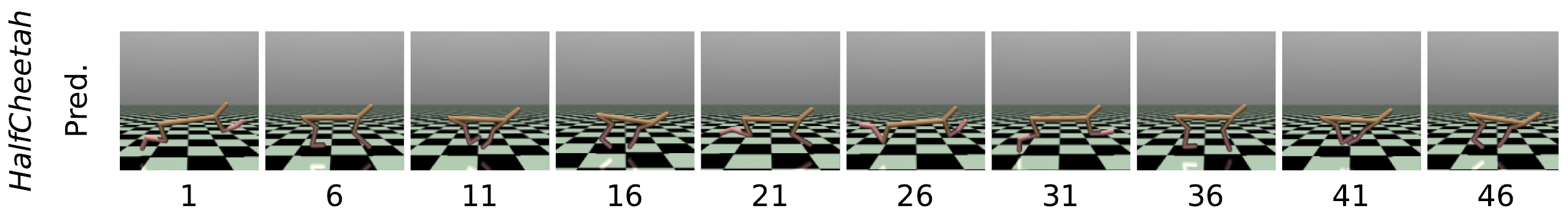}};
\node[rotate=90] at ([xshift=-0.4cm]img1.west) {ID};
% ----- Second row -----
\end{tikzpicture}
\hrule
\vspace{0.5mm}
\begin{tikzpicture}
\node[anchor=south west, inner sep=0] (img2) at (0,10)
    {\includegraphics[width=13cm]{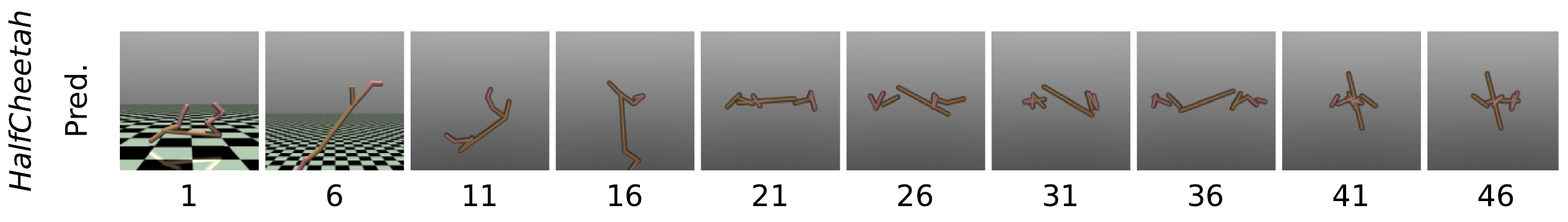}};
\node[rotate=90] at ([xshift=-0.4cm]img2.west) {OOD};
\end{tikzpicture}
\caption{Comparison of predicted rollouts in ID and OOD settings for Infoprop's PE.}
\label{fig:phys_traj_infoprop}
\end{figure}

Fig.~\ref{fig:phys_traj_infoprop} visualizes the proprioceptive state rollouts predicted by Infoprop's PE, when simulated in the environment, corresponding to the exemplary ID and OOD cases discussed in the main text.  

\begin{figure}[htb!]
\centering
\begin{tikzpicture}
% ----- First row -----
\node[anchor=south west, inner sep=0] (img1) at (0,0)
    {\includegraphics[width=11.5cm]{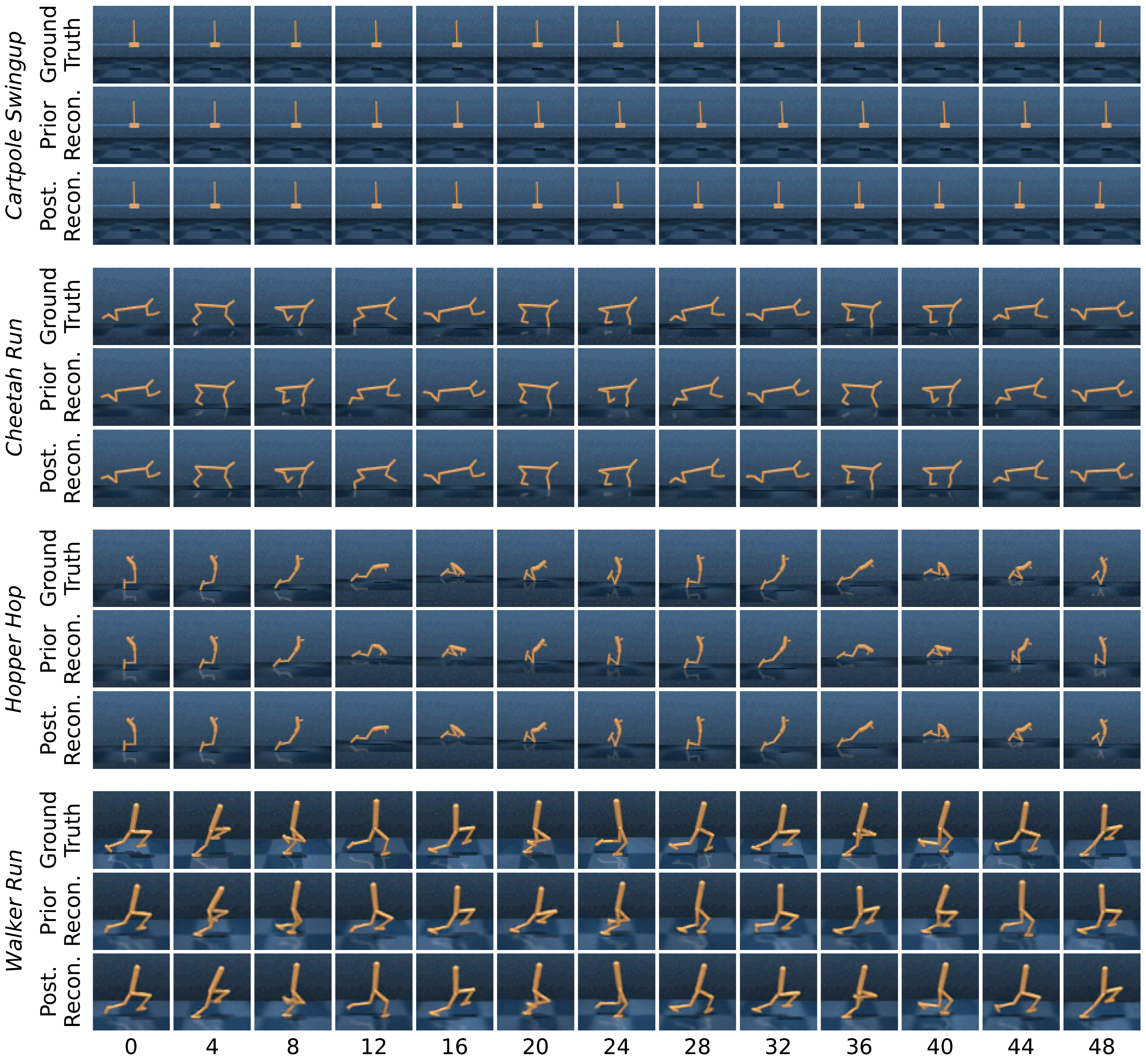}};
\node[rotate=90] at ([xshift=-0.4cm]img1.west) {RSSM};
% ----- Second row -----
\end{tikzpicture}
\hrule
\vspace{0.5mm}
\begin{tikzpicture}
\node[anchor=south west, inner sep=0] (img2) at (0,10)
    {\includegraphics[width=11.5cm]{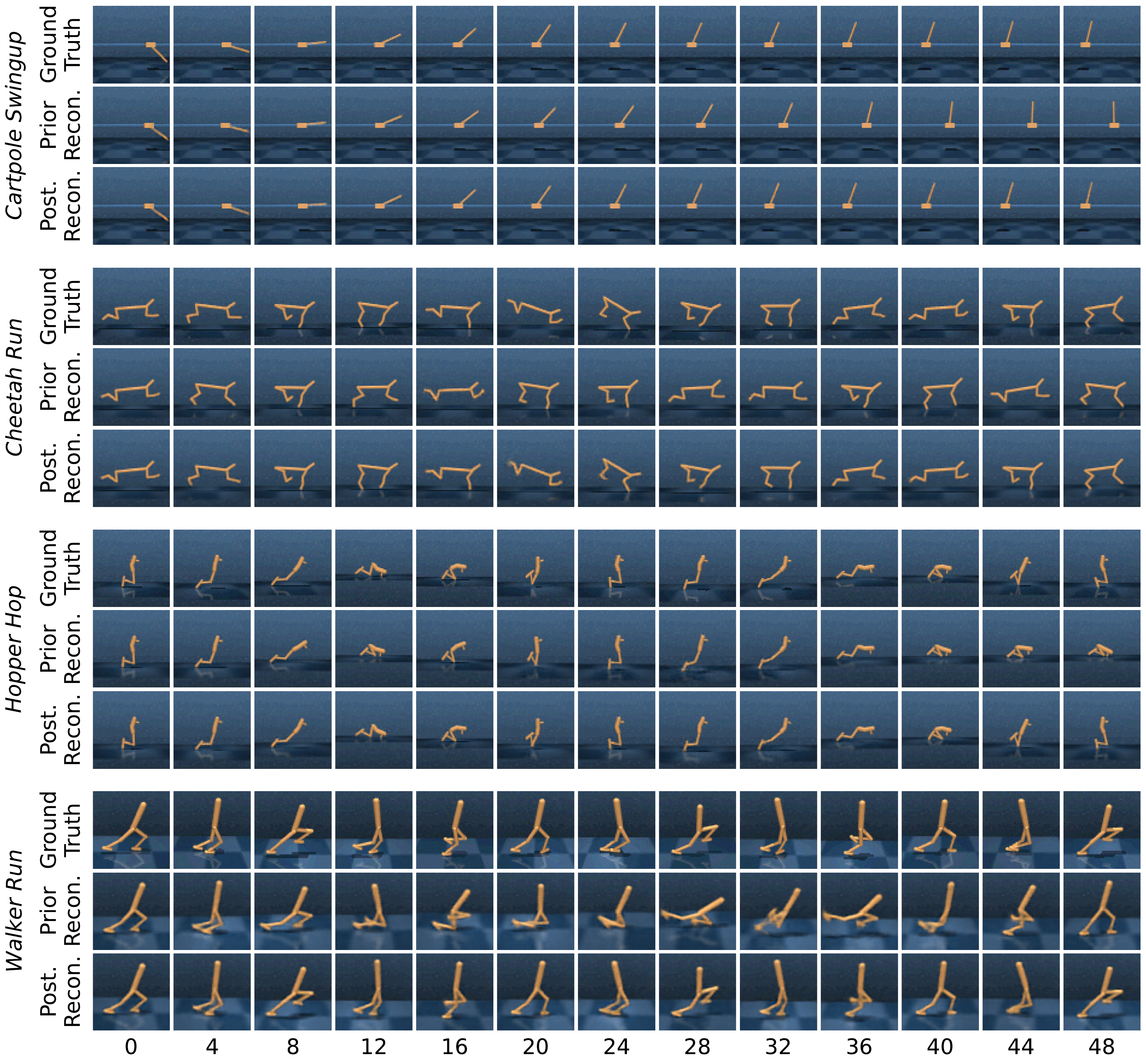}};
\node[rotate=90] at ([xshift=-0.4cm]img2.west) {Cat-RSSM};
\end{tikzpicture}
\caption{Step-wise comparison of reconstructed posterior-informed and closed-loop prior image rollouts in ID setting for RSSM and Cat-RSSM.}
\label{fig:id_img_recon}
\end{figure}

\begin{figure}[htb!]
\centering
\begin{tikzpicture}
% ----- First row -----
\node[anchor=south west, inner sep=0] (img1) at (0,0)
    {\includegraphics[width=11.5cm]{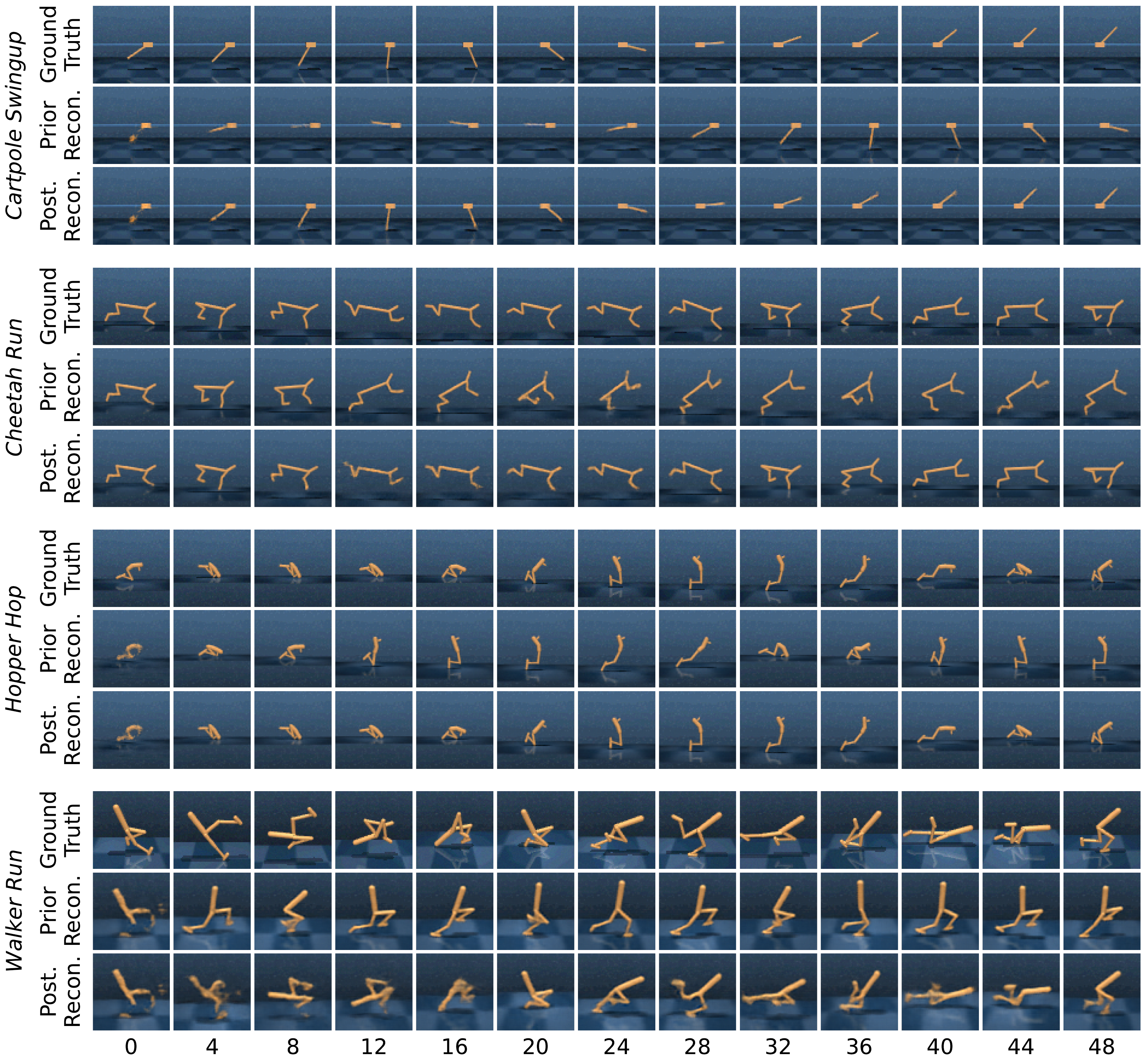}};
\node[rotate=90] at ([xshift=-0.4cm]img1.west) {RSSM};
% ----- Second row -----
\end{tikzpicture}
\hrule
\vspace{0.5mm}
\begin{tikzpicture}
\node[anchor=south west, inner sep=0] (img2) at (0,10)
    {\includegraphics[width=11.5cm]{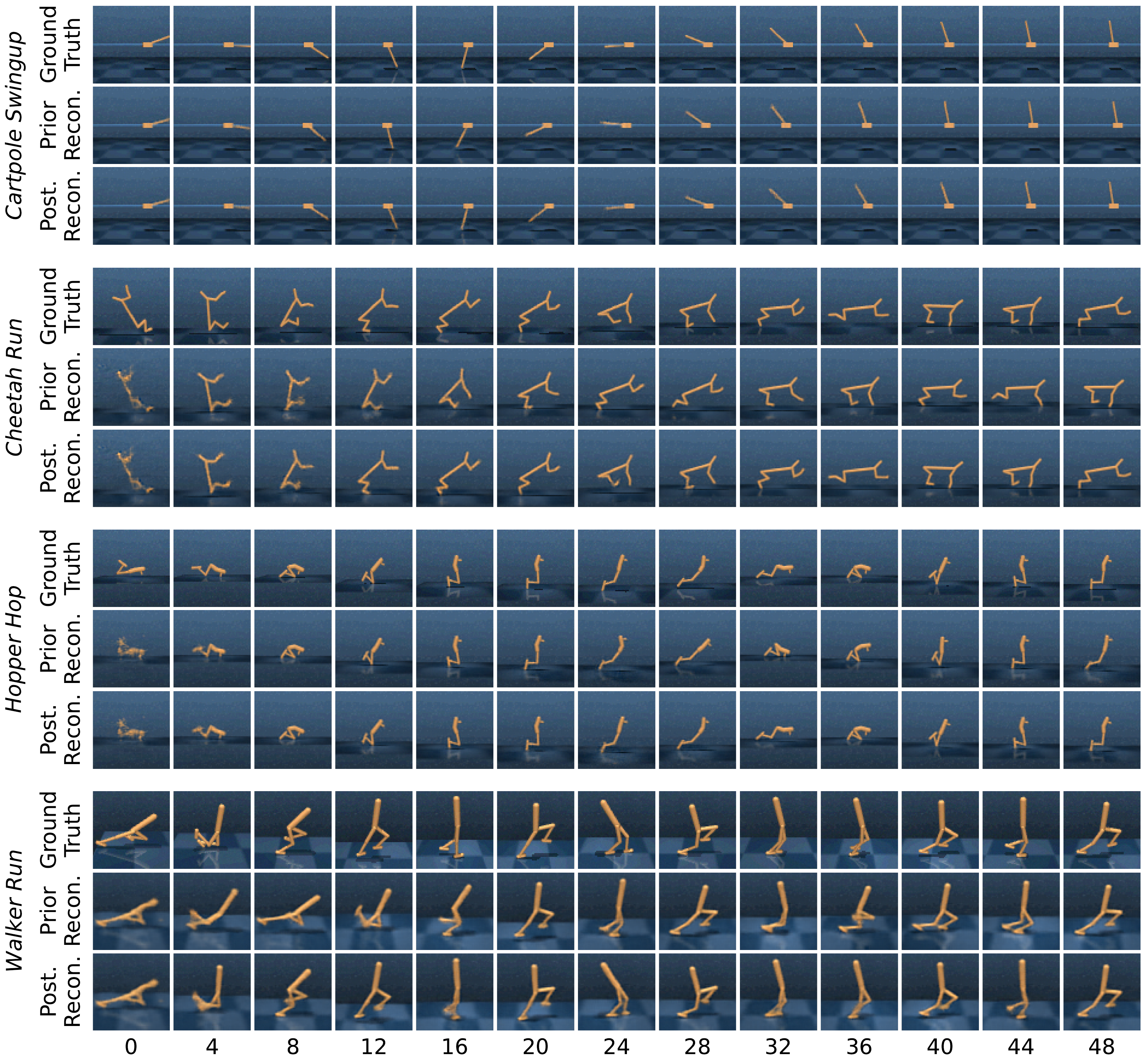}};
\node[rotate=90] at ([xshift=-0.4cm]img2.west) {Cat-RSSM};
\end{tikzpicture}
\caption{Step-wise comparison of reconstructed posterior-informed and closed-loop prior image rollouts in OOD setting for RSSM and Cat-RSSM.}
\label{fig:ood_img_recon}
\end{figure}

\begin{figure}[htb!]
\centering
\begin{tikzpicture}
% ----- First row -----
\node[anchor=south west, inner sep=0] (img1) at (0,0)
    {\includegraphics[width=11.5cm]{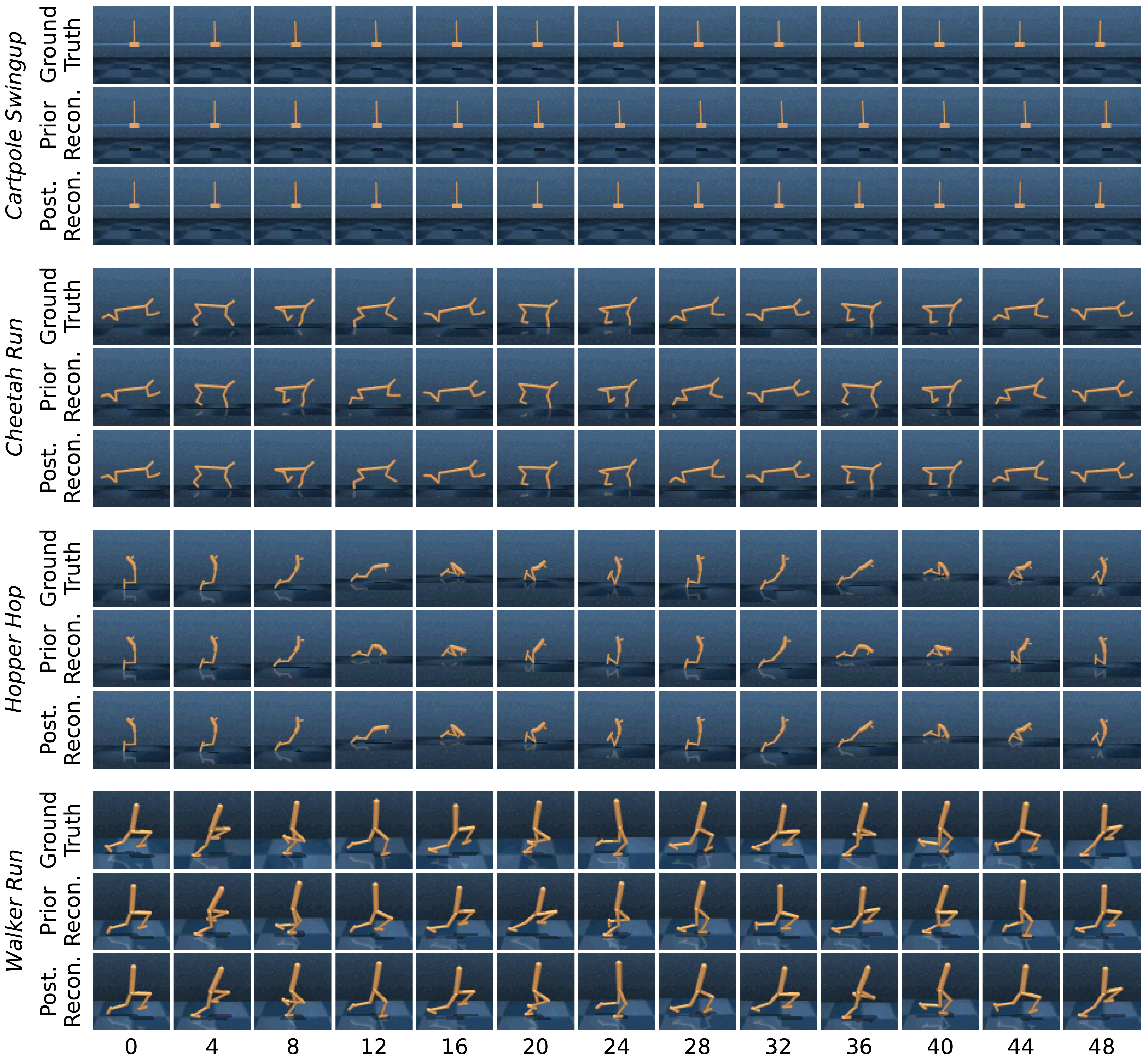}};
\node[rotate=90] at ([xshift=-0.4cm]img1.west) {RSSM};
% ----- Second row -----
\end{tikzpicture}
\hrule
\vspace{0.5mm}
\begin{tikzpicture}
\node[anchor=south west, inner sep=0] (img2) at (0,10)
    {\includegraphics[width=11.5cm]{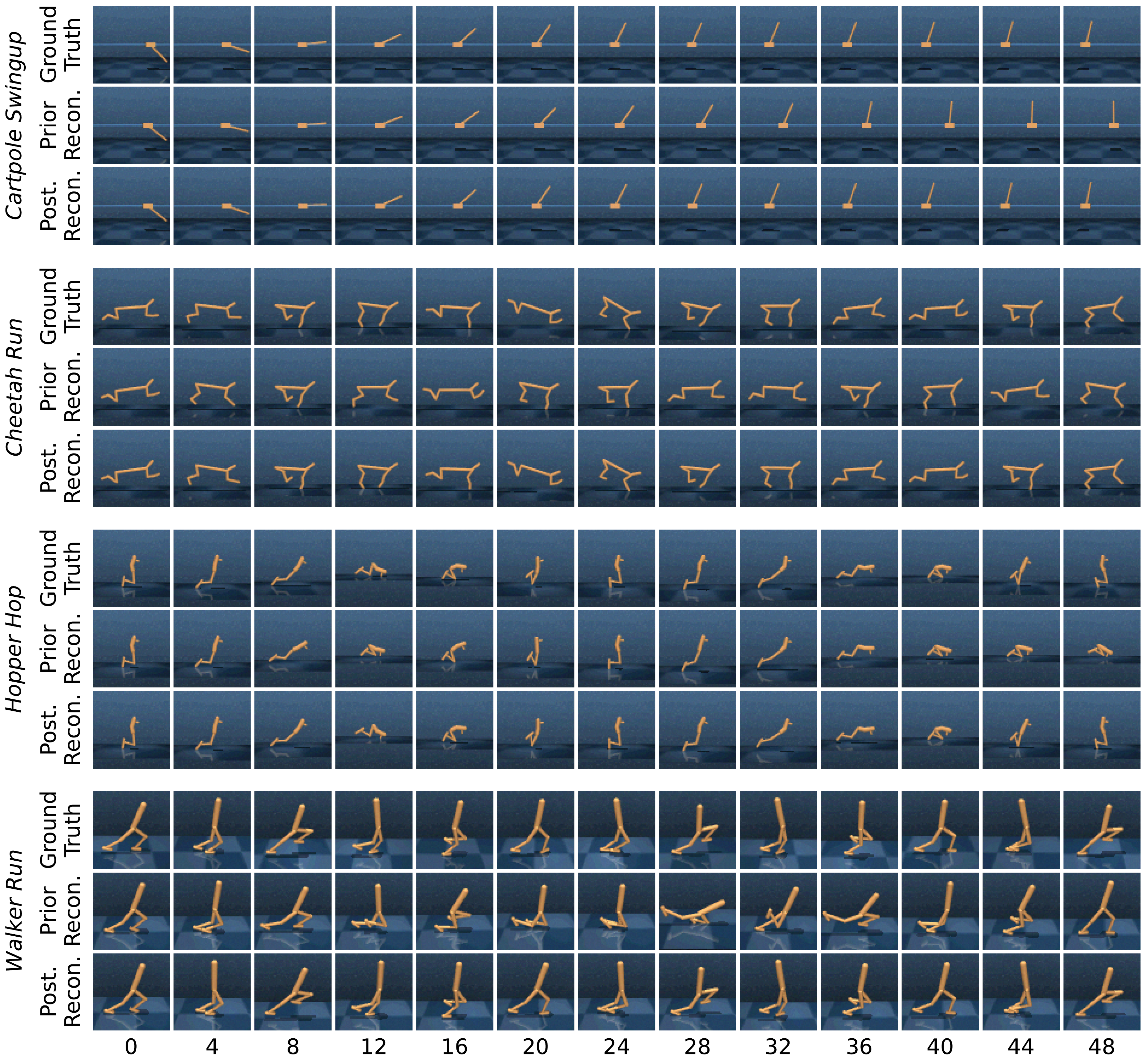}};
\node[rotate=90] at ([xshift=-0.4cm]img2.west) {Cat-RSSM};
\end{tikzpicture}
\caption{Step-wise comparison of reconstructed posterior-informed and closed-loop prior proprioceptive rollouts in ID setting for RSSM and Cat-RSSM.}
\label{fig:id_phys_recon}
\end{figure}

\begin{figure}[htb!]
\centering
\begin{tikzpicture}
% ----- First row -----
\node[anchor=south west, inner sep=0] (img1) at (0,0)
    {\includegraphics[width=11.5cm]{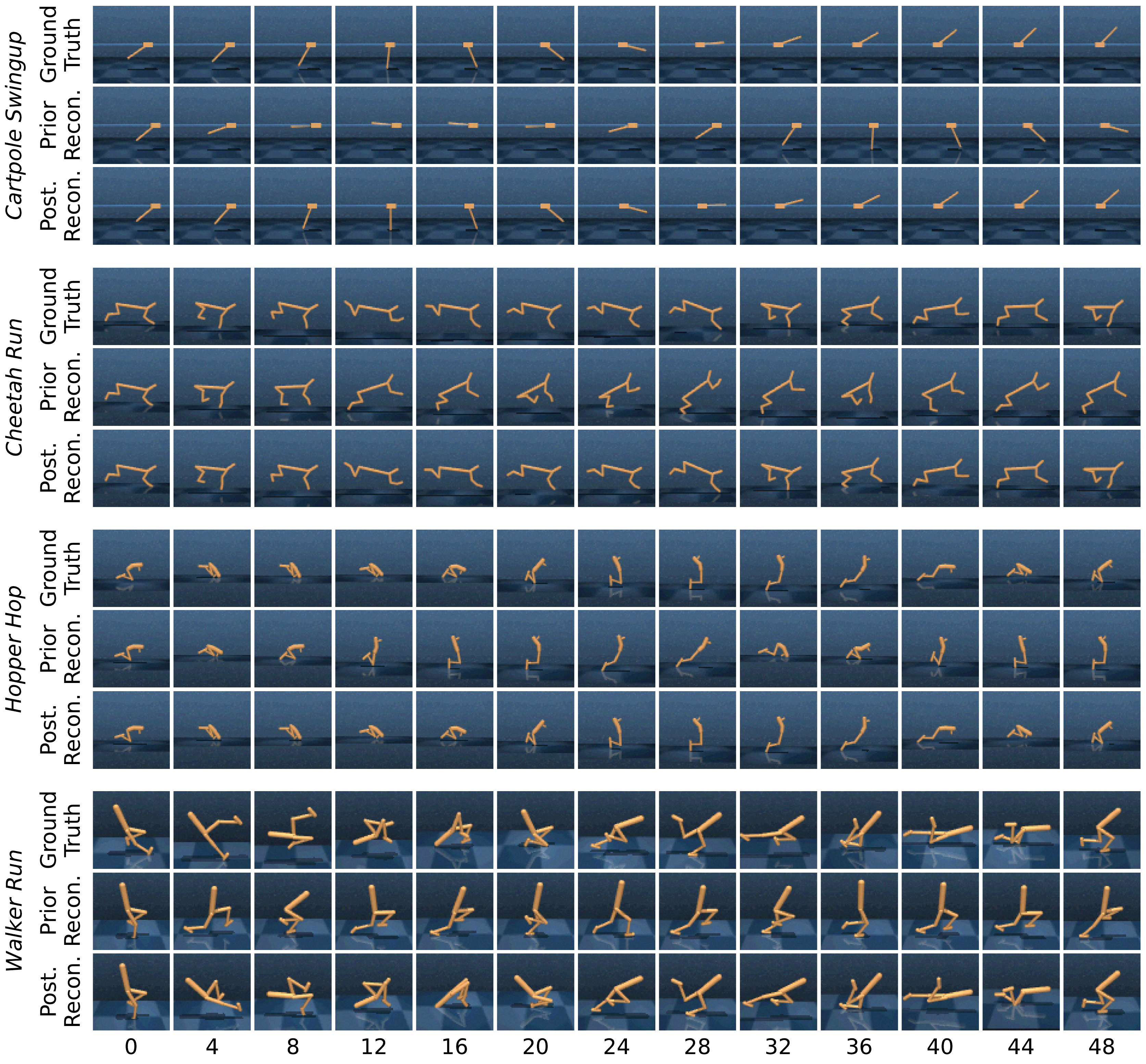}};
\node[rotate=90] at ([xshift=-0.4cm]img1.west) {RSSM};
% ----- Second row -----
\end{tikzpicture}
\hrule
\vspace{0.5mm}
\begin{tikzpicture}
\node[anchor=south west, inner sep=0] (img2) at (0,10)
    {\includegraphics[width=11.5cm]{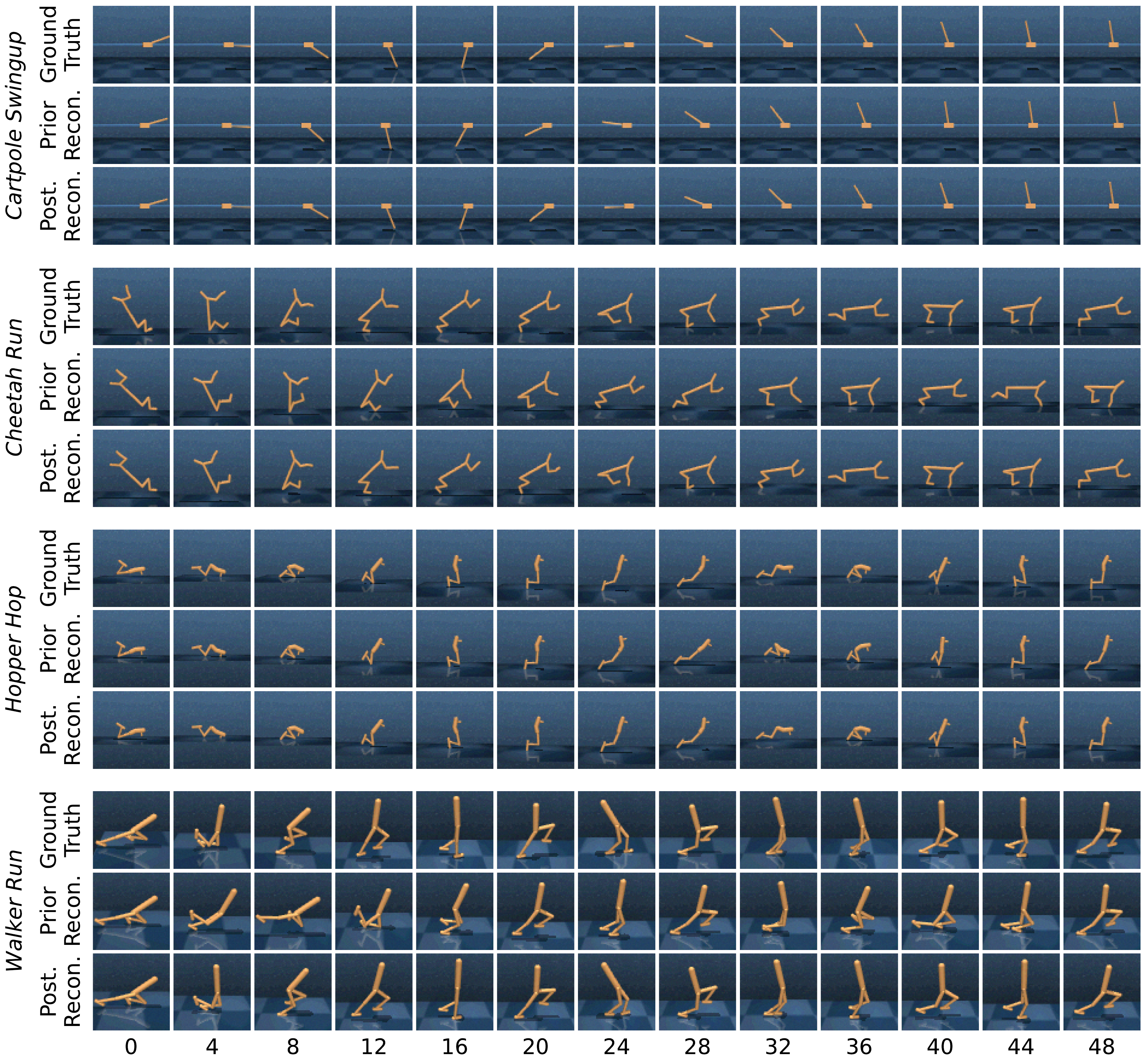}};
\node[rotate=90] at ([xshift=-0.4cm]img2.west) {Cat-RSSM};
\end{tikzpicture}
\caption{Step-wise comparison of reconstructed posterior-informed and closed-loop prior proprioceptive rollouts in OOD setting for RSSM and Cat-RSSM.}
\label{fig:ood_phys_recon}
\end{figure}

Figs.~\ref{fig:id_img_recon},~\ref{fig:ood_img_recon} and Figs.~\ref{fig:id_phys_recon},~\ref{fig:ood_phys_recon} compare simulator ground truth with exemplary image and proprioceptive rollouts, respectively, reconstructed from the corresponding posterior-informed and closed-loop prior rollouts. We visualize the first posterior-informed and closed-loop prior rollouts from our collection. Overall, the reconstructed image and proprioceptive state rollouts are largely consistent with each other, indicating that the proprioceptive decoder achieves reconstruction quality comparable to the image decoder. Posterior-informed rollouts remain close to the ground truth, whereas closed-loop prior rollouts exhibit increasing misalignment between predicted states and environment dynamics over longer horizons, independent of initial state setting (ID/OOD).

% --------- INFOPROP ROLLOUTS ---------

% --------- IMG ROLLOUTS ---------

% --------- PHYS ROLLOUTS ---------

\end{document}